\documentclass[applsci,article,preprints,pdftex,moreauthors]{Definitions/mdpi} 
\usepackage{microtype}

\newcommand{\fref}[1]{Fig.~\ref{#1}}
\newcommand{\sref}[1]{Section~\ref{#1}}
\newcommand{\tref}[1]{Table~\ref{#1}}

\usepackage[normalem]{ulem}
\usepackage{graphicx}
\usepackage{amsmath}
\usepackage{algorithm}
\usepackage[noend]{algpseudocode}
\usepackage{multirow}
\usepackage{bbding}

\usepackage{amsfonts}
\usepackage{amsmath}

\usepackage{setspace} 
\usepackage{helvet} 

\usepackage[labelformat=simple]{subcaption}

\DeclareCaptionLabelFormat{subcaptionlabel}{\normalfont(\textbf{#2}\normalfont)}
\usepackage{tabularx, booktabs}
\usepackage{indentfirst}
\usepackage{mathtools}

\newcommand{\covid}{COVID-19}

\usepackage{color}

\usepackage{soul}

\firstpage{1} 
\pubvolume{1}
\issuenum{1}
\articlenumber{0}
\pubyear{2022}
\copyrightyear{2022}
\externaleditor{Academic Editor: } 
\datereceived{6 July 2022}
\daterevised{} 
\dateaccepted{15 August 2022} 
\datepublished{} 
\hreflink{https://doi.org/} 

\Title{{Holistic} 
 Interpretation of Public Scenes Using Computer Vision and Temporal Graphs to Identify Social Distancing Violations}

\TitleCitation{Holistic Interpretation of Public Scenes Using Computer Vision and Temporal Graphs to Identify Social Distancing Violations}

\Author{{Gihan Jayatilaka} $^{1,2,\dagger}$\orcidA{}, Jameel Hassan $^{1,\dagger}$\orcidB{}, Suren Sritharan $^{3,\dagger}$\orcidC{}, {Janith Bandara Senanayaka} $^{1}$, Harshana~Weligampola $^{1,*}$\orcidD{}, Roshan Godaliyadda $^{1}$\orcidE{}, {Parakrama Ekanayake} $^{1}$, Vijitha Herath $^{1}$\orcidF{}, Janaka~Ekanayake $^{1,4}$\orcidG{} and Samath Dharmaratne $^{5}$\orcidH{}}

\AuthorNames{Gihan Jayatilaka, Jameel Hassan, Suren Sritharan, Janith Bandara Senanayaka, Harshana Weligampola, Roshan Godaliyadda, Parakrama Ekanayake, Vijitha Herath, Janaka Ekanayake and Samath Dharmaratne}

\AuthorCitation{Jayatilaka, G.; Hassan, J.; Sritharan, S.; Senanayaka, J.B.; Weligampola, H.; Godaliyadda, R.; Ekanayake, P.; Herath, V.; Ekanayake, J.; Dharmaratne, S.}

\address{%
$^{1}$ \quad Department of Electrical and Electronic Engineering, University of Peradeniya, {Peradeniya 20400, Sri Lanka}; {gihan@cs.umd.edu} (G.J.); jameel.hassan.2014@eng.pdn.ac.lk (J.H.); janith.b.senanayaka@eng.pdn.ac.lk (J.B.S.); roshangodd@ee.pdn.ac.lk~(R.G.); mpb.ekanayake@ee.pdn.ac.lk (P.E.); vijitha@ee.pdn.ac.lk (V.H.); ekanayakej@eng.pdn.ac.lk (J.E.)\\
$^{2}$ \quad Department of Computer Science, University of Maryland, {College Park, MD 20742, USA}\\
$^{3}$ \quad Department of Informatics, Technical University of Munich, {85748 Garching, Germany}; suren.sri@eng.pdn.ac.lk\\
$^{4}$ \quad School of Engineering, Cardiff University, {Cardiff, Wales, CF24 3AA, UK}\\ 
$^{5}$ \quad Department of Community Medicine, University of Peradeniya, {Peradeniya 20400, Sri Lanka}; samath.dharmaratne@mbbs.md
}

\corres{Correspondence: harshana.w@eng.pdn.ac.lk}

\firstnote{These authors contributed equally to this work.} 

\abstract{Social distancing measures are proposed as the~primary strategy to curb the~spread of the~\covid\ pandemic. Therefore, identifying situations where these protocols are violated has implications for curtailing the~spread of the~disease and promoting a~sustainable lifestyle. This~paper proposes a~novel computer vision-based system to analyze CCTV footage to provide a~threat level assessment of \covid\ spread. The~system strives to holistically interpret the~information in~CCTV footage spanning multiple frames to recognize instances of various violations of social distancing protocols, across time and space, as~well as identification of group behaviors. This~functionality is achieved primarily by utilizing a~temporal graph-based structure to represent the~information of the~CCTV footage and a~strategy to holistically interpret the~graph and quantify the~threat level of the~given scene. The~individual components are evaluated in~a~range of scenarios, and the~complete system is tested against human expert opinion. The~results reflect the~dependence of the~threat level on people, their physical proximity, interactions, protective clothing, and~group dynamics, with a~system performance of 76\% accuracy.
}


\begin{document}

\section{Introduction}
\label{sec:introduction}

\covid\ is a~viral infection that causes a~wide range of complications, primarily in~the~respiratory system~\cite{covid-pneumonia} along with other systems~\cite{covid-cardiovascular,covid-neuro}. As~per current statistics, even though the~virus has~a~comparatively small case mortality rate, it~has amassed a~massive fatality count due to its high infectiousness. The~World Health Organization (WHO) estimates that the~virus has infected around 338 million people and claimed more than 5.72 million lives as of December 2021. Despite the~availability of effective vaccines against virus spread, medical complications, and~mortality, complete global vaccination coverage is still far overdue. Furthermore, emerging variants cast some non-trivial obstacles to vaccine efficiency~\cite{delta-vaccine-efficacy,epsilon-vaccine-efficacy,vac_efficacy, vac_eff_mdpi}. Therefore, mitigating the~spread of the~disease through social distancing, mask wearing, hand washing, sanitizing, and~other practices of hygiene still remains indispensable by and large~\cite{hand-washing, gondor} to restore normalcy whilst ensuring the~safety of~health.

People, being a~social species, tend to exhibit group behaviors frequently. Therefore, even the~most mindful persons may violate social distancing protocols occasionally~\cite{SCSsocialvulnerability,SCSphysical-space-opt}. Even such occasional violation of social distancing protocols may garner a~risk of contracting \covid\ depending on~the~proximity or duration of the~violation~\cite{SCSphysical-barrier, SCSoffice-threat}. Conversely, monitoring such violations of social distancing protocols (i.e., proximity, duration as well as the~intensity of sudden events such as maskless cough or sneezing) provide vital tools for contact tracing, monitoring, and~eventually pandemic control. In~essence, observing social distancing protocol violations is a~task with many caveats. Thus, automating this manual process needs meticulous analysis~\cite{SCSreview-contacttracing}. 
The main two avenues of research have been (a) intrusive solutions where people are actively contributing to the~measurement (by handheld devices, etc.) and (b) non-intrusive solutions with zero burden on~the~people (which could be deployed to any situation irrespective of who is being monitored).

The first type (intrusive techniques) requires a~signal to be transmitted by the~people being tracked; i.e.,~methods of this type require an~active beacon by each tracked person. Such a~wearable device based on an~oscillating magnetic field for proximity sensing to monitor social distancing to prevent \covid\ has been presented in~\cite{bian2020wearable}. This~system was shown to be more robust than Bluetooth-sensing devices~\cite{bluetooth2020}, especially in~determining the~distance threshold limit. However, it~is practically difficult to deploy a~solution of this type in~a~public space in~a~real-world situation. Thus, a~non-intrusive solution is preferable for large-scale deployment in~public spaces as the~people who are being tracked are done so~passively.

Research in~non-intrusive techniques to monitor social distancing has led to a~large body of work utilizing computer vision techniques. The~major sub-tasks in~those approaches are the detection and tracking of people, and the~state of the~art for these sub-tasks is now primarily dominated by convolution neural networks (CNNs). Most recent applications combine YOLO~\cite{yolov2} and deepSORT~\cite{deepsort-2017} to form powerful tools which can achieve object detection and tracking in~real time, and it~is used to tackle object recognition problems in~different scenarios such as license plate recognition~\cite{imageNvision-license-plate}, road marking detection~\cite{imageNvision-road-marking}, pedestrian detection~\cite{imageNvision-pedestrian-detection},~agricultural production~\cite{imageNvision-apple-flower}, etc.\par 

The work in~\cite{ansari2021monitoring} is an~example of a~CNN framework built on~the~aforementioned detection and localization algorithms to detect people, calculate the~Euclidean distance between them and spot social distancing violations. A~similar approach using YOLOv3 is performed in~\cite{SCStopviewDL,ahmed2021deep} for birds-eye view (overhead) camera footage. However, such overhead viewpoints are not practically deployable in~public settings. An~SSD-based model is presented in~\cite{qin2021reaserch}, which also performs person detection and social distancing violation identification. The~performance is compared for each of the~deep learning models Faster RCNN, SSD, and~YOLO. Reference~\cite{rahim2021monitoring} utilizes the~YOLOv4 model for people detection in~low light instances to enforce social distancing measures. In~\cite{su2021novel}, a~spatio-temporal trajectory-based social distancing measurement and analysis method is proposed. This~problem has been further examined in~\cite{deepsocial-mdpi-applied-sci,vision-based-social-distancing-critical-density,social-distancing-fine-tuned-yolo-deepsort}.

While various solutions proposed in~the~literature strive to assess the~adherence to social distancing protocols, they fall short of incorporating factors such as mask wearing, which is critical to the~current \covid\ pandemic. The~presence or absence of a~mask on a~person greatly affects the~efficacy of the~social distancing protocols~\cite{wearing-masks}. Similarly, interperson interactions such as hugs, kisses, and~handshakes are more severe concerns than mere distancing amongst individuals~\cite{touch-spread-1,touch-spread-2} as far as the~person-to-person spreading of \covid\ is concerned. 
The detection of mask-wearing~\cite{Moxa3k,TinyMask,SCSmask-detection, imageNvision-facemask, HandsOff} as~well as the~detection of dyadic interactions~\cite{shinde2018yolo,sefidgar2015discriminative,van2018hands} has been explored in~computer vision as isolated and distinct problems.
However, to~the best of the~knowledge of the~authors, those factors have not been incorporated into a~unified and holistic solution for detecting violations of social distancing protocols in~the~literature \tref{tab:social_dist_sys}. Ignoring such factors vastly undermines the~robustness of vision-based techniques to tackle the~social distancing problem of \covid. \par

\begin{table}[H]
\caption{{Different} social distancing measures and the~handling availability in~our proposed~system.}
\label{tab:social_dist_sys}
\newcolumntype{C}{>{\centering\arraybackslash}X}
\begin{tabularx}{\textwidth}{>{\centering}m{4cm}>{\centering}m{5.1cm}C}
\toprule
\textbf{ Social Distancing Measure}   & \textbf{Specifics}                          & \textbf{Handled in~Our System} \\ \midrule
Physical distancing~\cite{jones2020two}     & Singapore (1 m), South Korea (1.4 m) & \checkmark                     \\
Mask wearing~\cite{kwon2021association}           & Practiced in~most of the~countries & \checkmark                     \\ 
Close contacts~\cite{durkin2021touch}         & Handshakes, hugging, etc.          & \checkmark                     \\ 
Hygiene practices~\cite{qian2020covid,kwon2021association}    & Washing hands, sanitizing, etc.    &                      \\ 
Restricted gathering~\cite{verani2020social,kwon2021association} & Indoor gatherings                 & \checkmark                     \\ \bottomrule
\end{tabularx}
\end{table}

In this light, the~system proposed in~this paper analyzes the~spatial and temporal interactions manifested over multiple frames.
A single frame was analyzed to recognize how people adhere to social distancing measures such as keeping proper distance, mask wearing, and~handshake interactions. 
The risk of spreading \covid\ increases when an~individual interacts with multiple people and~the nature of the~interaction.
On the~other hand, if~a certain set of people are in~a~``bubble'' and they remain so until the~end of observation, there is no change in~the~risk of spreading \covid. This~temporal analysis of identifying bio-bubbles is also included in~our proposed~model.

In this paper, the~design, implementation, and~testing of a~complete end-to-end system comprising of a~framework to fit in~different computer vision and deep learning-based techniques, a~representation to store the~output of the~deep learning models, and~an interpretation technique to evaluate the~threat level of a~given scene are discussed. The~key contributions of this paper are as follows:

\begin{itemize}
    \item A deep learning-based system to monitor social distancing violations and \covid\ threat parameters. The~system can utilize multiple computer vision modules to extract different information from the~video sequence such as the~number of people, their location, their physical interactions, and~whether they wear~masks. 

    \item A temporal graph representation to structurally store the~information extracted by the~computer vision modules. In~this representation, people are represented by nodes with time-varying properties for their location and behavior. The~edges between people represent the~interactions and social~groups.
    
    \item A methodology to interpret the~graph and quantify the~threat level in~every scene based on primary and secondary threat parameters such as individual behavior, proximity, and~group dynamics extracted from the~graph representation.
\end{itemize}

\section{Proposed~Solution}
\label{sec:methodology}

\begin{figure}[H]
    \includegraphics[width=0.95\textwidth]{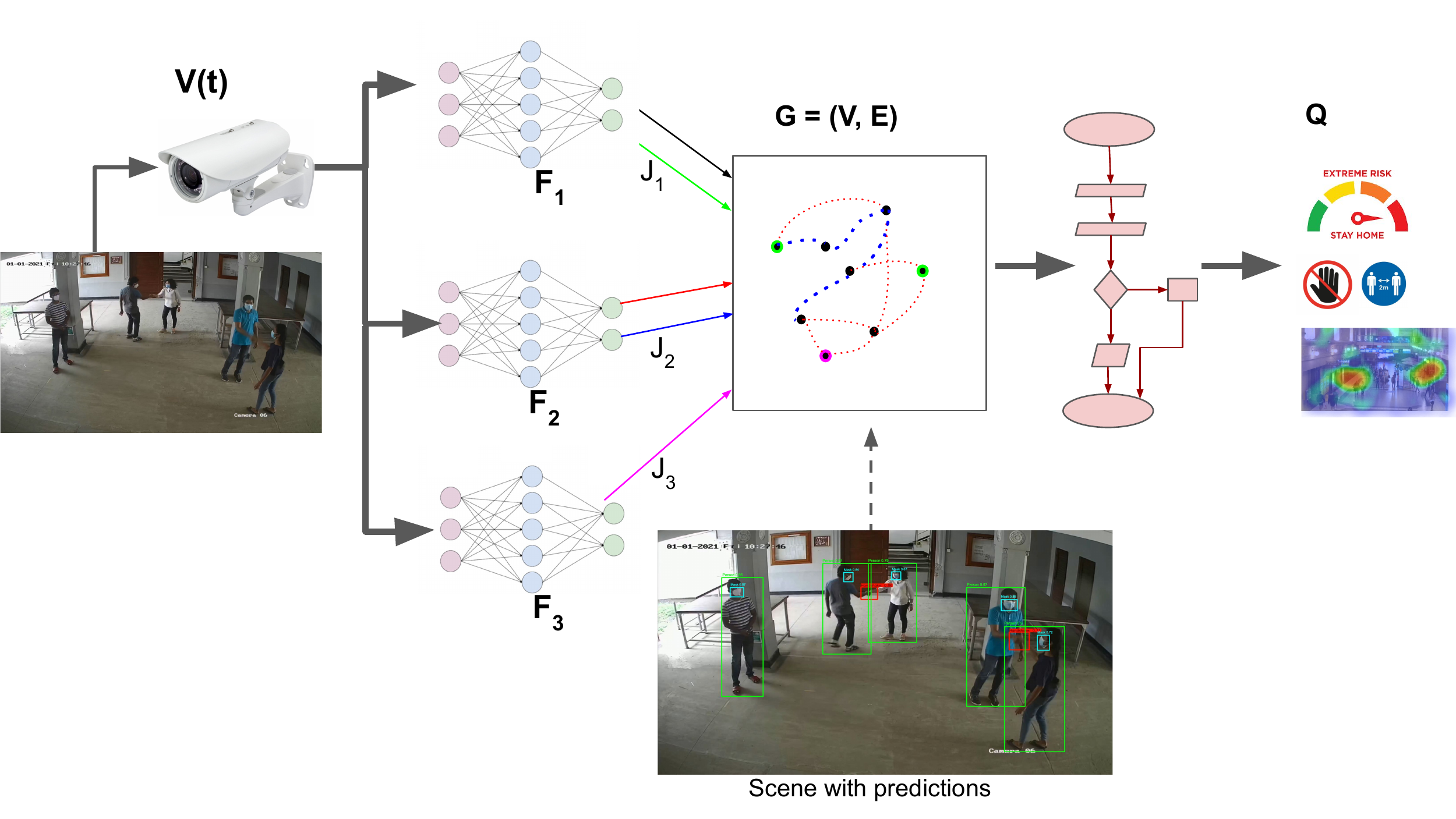}
    \caption{{A} 
 high-level overview of the~proposed~system. } 
    \label{fig:proposed-system}
\end{figure}

This section explains the~graph-based computer vision framework proposed to quantify the~risk of \covid\ transmission in~various public scenarios. The~input video feed from closed circuit television (CCTV) footage is first used to extract key information such as people, handshake interactions, and~face masks through computer vision models. The~proposed system then quantifies the~risk of transmission of \covid\ by encoding the~extracted information into a~temporal graph and interpreting it~using a~function for the~threat of transmission developed in~this paper. An~overview of the~proposed system is depicted in~Figure~\ref{fig:proposed-system}.\par

The system takes a~video stream $V_{in}(t)$ as the~input, where $t$ denotes the~frame number. The~video stream is considered to be captured from a~CCTV system camera mounted at the~desired vantage point with a~known frame rate. $V_{in}(t)$ is a~three-dimensional matrix with the~dimensions $H \times W \times 3$, where $H$ and $W$ denote the~frame's height and width,~respectively.

The video feed, $V_{in}(t)$, was passed into a~series of functions $F_i\:;\:i \in \{p, d, g, h, m\}$. Each $F_i$ processes a~video frame and produces different information as
\begin{equation}
    J_{i}(t) = F_i \big( V_{in}(t) \big)
\end{equation}
where $J_{i}(t)$ denotes an~output, such as the~locations of people, handshake interactions, or~the presence of face masks. While the~functions $F_i$s process individual frames, processing a~sequence of frames is required to analyze this information across time. Therefore, a~collection of trackers $\overline{F_i}$ was employed to track the~above-mentioned detections provided by $F_i$s over time as
\begin{equation}
    \label{eqn:stateful-fun}
    [S_i(t), \overline{J_i}(t)] = \overline{F_i}{} \Big( V_{in}(t), J_{i}(t), S_i(t-1) \Big)
\end{equation}
\textls[-35]{where $S_i(t)$ is the~state and $\overline{J_i}(t)$ is the~tracking interpretations based on~the~sequential~information.}\par

The list of functions utilized to obtain spatial information necessary for detecting and localizing persons, interactions, and~face masks follows:
\begin{enumerate}
    \item People detection ($F_p$) and tracking ($\overline{F}_p$).
    \item Distance estimation ($F_d$) and group identification ($F_g$).
    \item Identifying and localizing physical interaction (handshakes) ($F_h$).
    \item Mask detection ($F_m$).
\end{enumerate}

The information retrieved by the~aforementioned functions, which is critical for calculating the~social distancing violation measure, was encoded in~a~graph $G=(V,E)$. Sections~\ref{sec:people-tracking}--\ref{sec:mask} define the~functionality of each system component that works together to populate the~graph $G$, while \sref{sec:graph-representation} offers a~full explanation of the~data contained in~the~graph.
Finally, graph $G$ was interpreted in~the~manner described in~\sref{sec:threat-quantification} in~order to provide actionable insights based on~the~threat level analysis of the~analyzed video. For~ease of understanding, the~notations used in~this work are listed in~a~table in~{Abbreviations}.

\subsection{People Detection and~Tracking}
\label{sec:people-tracking}
This section discusses the~proposed framework's people detection and tracking models. The~people in~the~scene were detected using the~$F_p$ detection model and then tracked over different frames using the~$\bar{F_p}$ tracking model. The~detection model used for this purpose provides a~bounding box for the~person's position, whilst the~tracking model assigns each person a~unique ID and tracks them through~time.

\begin{figure}[H]
    \includegraphics[width=8cm, height=9cm]{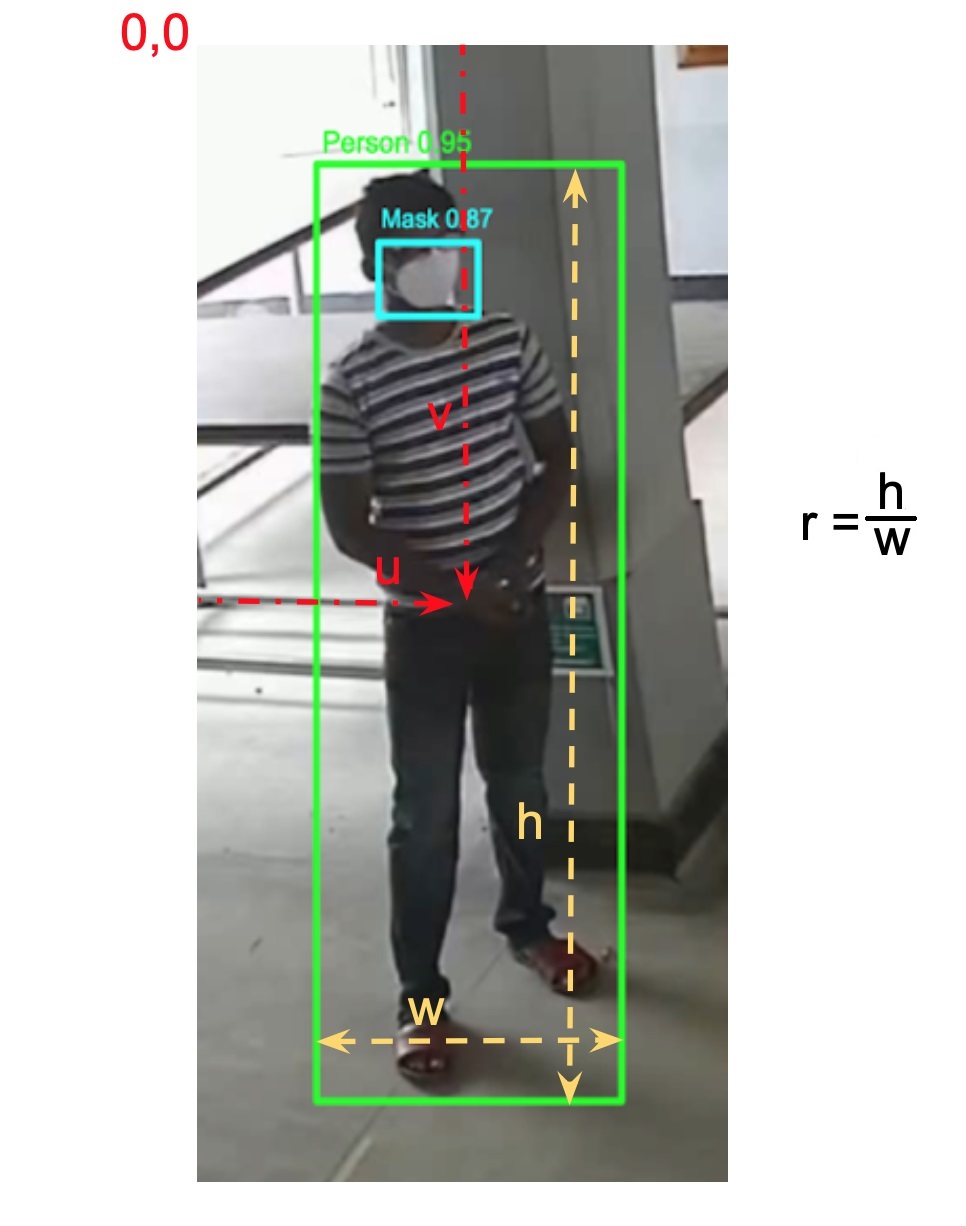}
    \caption{The parameters for the~bounding~boxes.} 
    \label{fig:BB-explained}
\end{figure}

The detection model provides a~time-varying vector containing information on people's spatial location. It~is defined as $J_p(t) = \{bb_{p1}(t), bb_{p2}(t), \ldots, bb_{pk}(t), \ldots, bb_{pn}(t)\}$, where $n$ is the~number of bounding boxes and $bb_{pk}{(t)} = (u, v, r, h, c_p)$ is a~five-tuple that represents the~bounding box representing a~person at time $t$. In~$bb_{pk}{(t)}$, variables $u$ and $v$ represent the~two-dimensional coordinates of the~bounding box's center, $r$ represents the~bounding box's aspect ratio, $h$ represents the~bounding box's height, and~$c_p$ represents the~detection's confidence level, as shown in~Figure~\ref{fig:BB-explained}. The~tracker assigns an~ID and updates bounding box information based on previous and current data. The output of the~tracker is defined as $\overline{J_p}(t) = \{bbi_{p1}(t), bbi_{p2}(t), \ldots, bbi_{pk}(t), \ldots, bbi_{pn}(t)\}$, where $bbi_{pk}{(t)} = (u, v, r, h, c_p, i)$ is a~six-tuple representing updated bounding box information with assigned ID, $i$, for~$k${th} person.

Given its robustness and real-time prediction capabilities, the~YOLO network~\cite{yolov4} for people detection ($F_p$) and the~DeepSORT algorithm~\cite{deepsort-2017} for tracking ($\bar{F_p}$) were used in~this paper. DeepSORT by itself handles minor occlusions for people by the~Kalman filtering mechanism. We~implement another level of interpolation to handle the~missed detections on top of this. Given an~image, the~YOLO network predicts the~bounding boxes of many predefined object classes that are present in~a~scene. Following that, the~output is created by applying non-max suppression~\cite{canny} and filtering the~bounding boxes belonging to people. The~DeepSORT algorithm then assigns indices, $\overline{J_p}$, to~these detected bounding boxes using the~Mahalanobis distance and the~cosine similarity of the~deep appearance descriptors. The~publicly available weights trained using the~COCO dataset~\cite{COCO} were used to initialize the~weights of the~YOLO model, whereas the~weights trained using the~MOT dataset~\cite{mot-dataset-2020} were used to initialize the~DeepSORT~model.

\subsection{Distance~Estimation}
\label{sec:distance-estimation}

This section discusses the~method for estimating the~distance between identified individuals. The~distance between people was estimated in~three steps: first, by~identifying the~people's standing locations in~the~video, then by performing perspective transform and finally by measuring their Euclidean distance~\cite{perspective-transform}.

\begin{figure}[H]
    \includegraphics[width=0.97\textwidth]{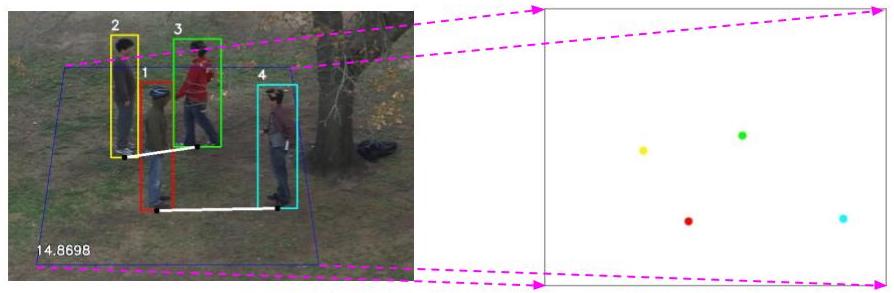}
    \caption{{Perspective transformation}. The~(\textbf{right}) frame is a~visualization of how a~camera-captured scene (\textbf{left}) is projected to the~'floor plane' after perspective transformation. The~trapezoidal floor is being transformed into a~square.}
    \label{fig:perspective-transform}
\end{figure}

First, the~standing locations of the~people $s_{(i,t)}$ (denoted by thick black dots in~Figure~\ref{fig:perspective-transform}) were determined using the~bounding box data as follows,
\begin{equation}
    \label{eqn:standing-locations}
    s_{(i,t)} = (u, v+ 0.5h).  
\end{equation}

The standing locations were then transformed via perspective transform from an~overhead wall mount camera viewpoint to a~two-dimensional bird's eye viewpoint. The~required transformation matrix $M_T$ was obtained as follows,
\begin{align}
\begin{split}
    R' &= M_TR\\
    R'R^T &= M_TRR^T\\
    R'R^T(RR^T) ^{-1} &= M_T (RR^T)(RR^T) ^{-1}\\
    M_T &= R'R^T(RR^T)^{-1}
\end{split}
\end{align}
where the $R$ values are $2\times4$ matrices that contain~the~coordinates of four reference points in~the~video frame (refer blue trapezoid in~Figure~\ref{fig:perspective-transform}---left) and the~corresponding coordinates of those four points in~the~two-dimensional plane. This~two-dimensional plane is referred to as the~``floor plane'' (refer Figure~\ref{fig:perspective-transform}---right). The~projections were performed as,
\begin{equation}
    floorLocation_{(i,t)} = M_T\ s_{(i,t)}   
\end{equation}
where $s_{(i,t)}$ are the~input coordinates from \eqref{eqn:standing-locations} and $floorLocation_{(i,t)}$ are the~output coordinates on~the~floor plane. Finally, the~distances between each pair of people $i$ and $j$ in~frame $t$ were calculated as
\begin{equation}
    dist_{(i,j,t)} = ||floorLocation_{(i,t)}-floorLocation_{(j,t)}||
\end{equation}

Since the~detected bounding boxes of people cannot be directly used to estimate distances between people due to the~overhead camera viewing angle, the~estimation is performed after perspective transform. The~transform is performed based on~the~following assumptions. These~assumptions hold for most of the~scenes with a~CCTV~camera. 

\begin{enumerate}
    \item All the~people are on~the~same plane.
    \item The camera is not a~fisheye-like camera.
    \item The camera is placed at an~overhead level.
\end{enumerate}
\vspace{-6pt}

\subsection{Group~Identification}
\label{sec:group-identification}
The group identification model discussed in~this section utilizes the~people detection, tracking and distance estimation models introduced in~Sections~\ref{sec:people-tracking} and \ref{sec:distance-estimation}. This~was achieved by two algorithms $F_g$ and $\overline{F}_g$. $F_g$ was run on~the~information from individual frames, while $\overline{F}_g$ analyzed the~results from $F_g$ across time to properly deduce which people fall into groups based on sustained physical~proximity.

Given a~frame $V_{in}(t)$, a~matrix $M_d(t)$ called the distance matrix is created based on~the~calculated distances between people. The~affinity matrix $M_a(t)$ was then calculated as follows,
\begin{equation}
    M_a = \exp{(-\alpha M_d)}
\end{equation}
where $\alpha$ is an~input parameter that is used to introduce the~camera and scene pair to a~scale. This~parameter acts as~a~threshold for the~closeness of people in~the~2D projected plane prior to clustering. Then, clustering was performed on $M_a$ to split the~people into clusters.
\begin{equation}
    clusters  = spectral\_clustering(M_a)
\end{equation}

According to the~group identification model, a~person is considered to be a~member of a~group if~they are close to at least one member of the~group.  While conventional clustering algorithms attempt to minimize the~distance between individual elements and the~cluster center, this is not how humans behave. As~a result, this result was obtained using spectral clustering of affinity matrices~\cite{spectral-clustering}. Human behavior, on~the other hand, cannot be analyzed in~terms of discrete frames. As~a result, a~temporal analysis of the~clusters was performed to determine the~actual groups of people using a~time threshold $\tau$. The~primary idea is that a~group is detected only if~it~persists for a~specified time period $\tau$.

People $P_i \in P$ were being clustered from a~video frame at time $t$ as follows,
\begin{equation}
    cluster\_id(P_i,t) \longleftarrow spectral\_clustering(M_a(t))\\
\end{equation}
\begin{equation}
    \label{eq:clusterid-same}
    cluster\_id(P_i,t) =  cluster\_id(P_j,t) \qquad \text{if} \quad \exists \quad t_0\ \text{s.t} \quad t_0 \leq t \leq t_0+\tau
\end{equation}
where $P_i$ and $P_j$ were considered to be in~the~same social group as per Equation~(\ref{eq:clusterid-same}). Social distancing violations between the~people in~the~same social group was ignored in~the~proposed system as justified in~\sref{sec:introduction}. For~cases involving a~few people, a~simplified algorithm based on naive thresholding of interpersonal distance violation occurrences was used instead of spectral clustering. When spectral clustering was used, $\tau$ was picked so that a~group should be in~proximity for 10 s. For~the thresholding case, people spending upwards of $20\%$ of the~time in~proximity were considered groups.

\subsection{Mask~Detection}
\label{sec:mask}

This section describes the~model used to detect the~presence/absence of masks. The~framework's mask recognition stage entails identifying and tracking the~presence (or absence) of masks. The~model used for this purpose computes the~bounding box of the~face as well as the~degree of confidence in~the~presence of a~mask. As~with prior object detection models, this model outputs a~time-varying vector representing the~spatial localization information for faces as $J_m(t) = \{bb_{m1}(t), bb_{m2}(t), \ldots, bb_{mk}(t), \ldots, bb_{mn}(t)\}$, where $n$ is the~number of face bounding boxes at time $t$ and $bb_{mk}{(t)} = (u, v, r, h, c m) $ is a~five-tuple representation of the~bounding box encompassing a~detected face at time $t$. The~variables $u, v$, $r$, and~$h$ have the~same definitions as those in~\sref{sec:people-tracking}. The~confidence measure $c_m = (c_{mask}, c_{nomask})$ is a~two-tuple in~which $c_{mask} \in [0, 1]$ indicates the~probability of the~presence of a~mask and $c_{nomask} \in [0, 1]$ indicates the~absence. Similar to \sref{sec:people-tracking}, the~tracking model returns a~vector of the~same size $\overline{J_m}(t)$ containing tracked bounding boxes for each $t$. \par

Similar to \sref{sec:people-tracking}, the~YOLO network was utilized for mask detection and the~DeepSORT algorithm was utilized for tracking the~masks across frames. The~YOLO model was first initialized with the~pre-trained COCO weights and then fine-tuned using the~images from the~Moxa3k dataset~\cite{Moxa3k} as well as the~UT and UOP datasets, which were labeled for mask detection. The~DeepSORT model used the~weights trained using the~MOT dataset~\cite{mot-dataset-2016} for initialization. The~DeepSORT algorithm handles minor occlusions of masks. However, another layer of interpolation (such as for people detection) was not implemented because the~algorithm is supposed to detect when people remove masks ({i.e.,} people cannot disappear while masks can). 

\subsection{Graph~Representation}
\label{sec:graph-representation}

\textls[-15]{The information extracted using different models in~Sections~\ref{sec:people-tracking}--\ref{sec:mask} need to be combined  to provide meaningful insights into the~threat level of the~given scene. This~is accomplished by encoding the~data into a~graph structure. This~section describes how the~graph structure is modeled using the~different outputs from the~models for~interpretation.}

\begin{figure}[H]
    \begin{subfigure}[]{0.45\linewidth}
        \centering
        \includegraphics[width=\textwidth]{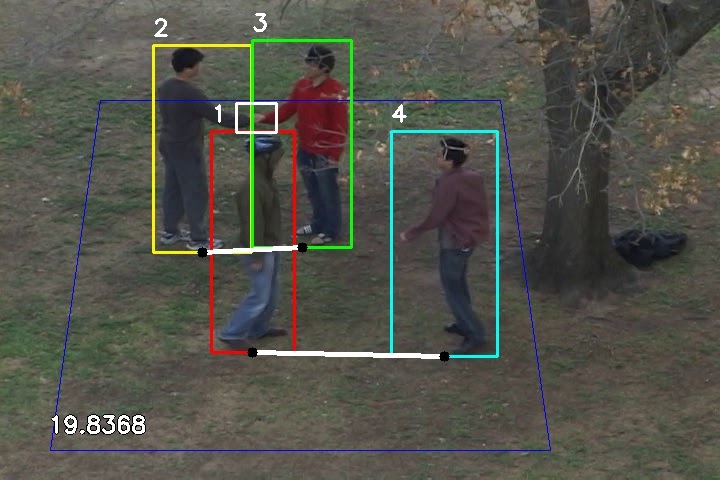}
\caption{\centering}
        \label{fig:graph_vid}
    \end{subfigure}
    \hfill
    \begin{subfigure}[]{0.45\linewidth}
        \centering
        \includegraphics[width=.88\textwidth]{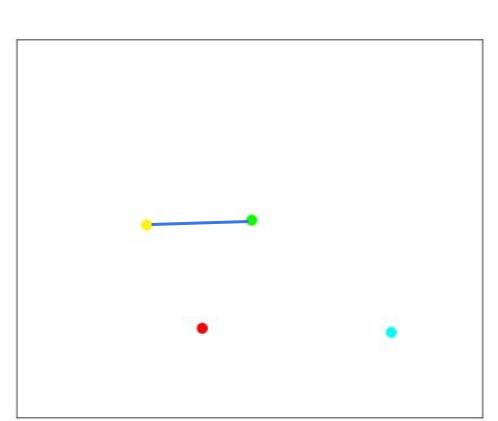}
\caption{\centering}
        \label{fig:graph_net}
    \end{subfigure}
    \caption{Graph representation figure. {(\textbf{a}) Bounding boxes for people and~handshake; (\textbf{b}) Corresponding graph~representation.}}
    \label{fig:graph-representation}
\end{figure}

\textls[-25]{The information retrieved from the~video is stored as~a~time-varying graph $G(t)$ given by}
\begin{align}
    G(t) &= \big(V(t),E(t) \big)
    \shortintertext{\centering and}
    V(t) &= \{v_1(t), v_2(t), \ldots, v_n(t)\}\\
    E(t) &= \{e_{1, 1}(t), e_{1, 2}(t), \ldots, e_{i, j}(t), \ldots, e_{n, n}(t)\}
\end{align}
where $V(t)$ is the~set of vertices and $E(t)$ is the~set of edges at time $t$. 
Each person $P_i$ is denoted by a~vertex $v_{i}(t)$ which contains the~features representing the~person extracted from the~video as time-varying vertex parameters. The~vertex $v_{i}(t)$ is given by
\begin{equation}
    v_{i}(t) = [\,location_i(t),\ mask_i(t),\ group_i(t)\,]
\end{equation}
where $location_i(t) = (x_i(t), y_i(t))$ is a~two-tuple that represents the~position of the~person $P_i$ at time $t$ obtained through perspective transform to a~bird's-eye view position on a~2D plane (refer to \sref{sec:distance-estimation}).
$mask_i(t) = c_m$ is two-tuple, which shows the~confidence level that a~person $P_i$ is wearing a~mask at time $t$. This~information is extracted from $bb_{mi}(t)$ depending on~the~index $ID_{mi}(t)$ (refer to \sref{sec:mask}). 
$group_i(t)$ is a~matrix that represents the~probability that two people belong to the~same group (refer to \sref{sec:group-identification}). The~edge $e_{i,j}(t)$ is a~binary value (0/1) that represents the~presence (denoted by 1) or absence (denoted by 0) of an~interaction between person $P_i$ and $P_j$ at time $t$ detected using~\cite{HandsOff}. $E(t)$ is stored as~a~sparsely filled adjacency matrix with null values for instances where interactions are not~detected. {A visual example} of a frame and its constructed graph is shown in \fref{fig:graph-representation}.

\subsection{Threat~Quantification}
\label{sec:threat-quantification}
The information extracted from the~models described in~the~proposed system \linebreak in~Sections~\ref{sec:people-tracking}--\ref{sec:graph-representation} needs to be processed from the~created temporal graph in~order to provide a~quantifiable metric that denotes the~risk of transmission for the~given scene/frame. In~this section, the~derivation of the~threat level function which quantifies the~threat of the~given frame is described in~detail. \par

\tref{tab:threatparams} contains a~list of the~parameters that contribute to the~spread of \covid. The~parameters are divided into two categories: primary and secondary parameters, which will be discussed further in~this section using the~threat level function. First, we~calculate the~threat level contribution of each pair of people in~the~frame at time $t$ as described in~\eqref{eqn:threat-general}. Then, we~find the~threat level of the~particular frame as per \eqref{eqn:threat-general-sum}.
\begin{equation}
    T(t) = \sum_{(v_1,v_2 \in V)}{T_{v_1,v_2}(t)}
    \label{eqn:threat-general-sum}
\end{equation}
\begin{equation}
    T_{v_1,v_2}(t) = \sum_{p_i \in \mathbb{P}}{p_i(v_1,v_2)} \times \prod_{q_j \in \mathbb{Q}}{\epsilon_j - q_j(v_1,v_2)}
    \label{eqn:threat-general}
\end{equation}

$\mathbb{P} = \{p_h, p_d \}$ is the~set of parameters that directly attributes to the~transmission of \covid\ from one person to another. This~includes the~distance between people and the~handshake interactions. As~the distance between people (people coming close) and their interactions (handshakes) play a~primary role in~the~\covid\ virus transmission, these values were first considered as the~primary parameters $\mathbb{P}$. The~probability of two people shaking hands $p_h$ and the~probability of them coming extremely close $p_d$ were represented as scalar values in~the~range $[0, 1]$, where 1 represents a~high probability of occurrence {(for the~distance probability, $1m$ is used as the~threshold distance for being extremely close in~this study)}.{} 

$\mathbb{Q} = \{q_m, q_g\}$ is the~set of secondary parameters which are relevant only when two people are in~close proximity, and~in such a~case, these parameters can increase or decrease the~probability of \covid\ transmission accordingly. This~includes whether people are wearing masks, since two people not wearing masks is irrelevant if~they are far apart, and~whether the~persons belong to the~same~group.

\vspace{-6pt}

\vspace{-6pt}
\begin{table}[H]
    \caption{Parameters used in~threat~quantification.}
\newcolumntype{C}{>{\centering\arraybackslash}X}
\begin{tabularx}{\textwidth}{>{\centering}m{3cm}>{\centering}m{3cm}C}
\toprule
\textbf{Set}       & \textbf{Notation} & \textbf{Description}               \\ \midrule
\multirow{2}{*}{$\mathbb{P}$} & $p_d$               & Distance between people            \\ 
                   & $p_h$               & Handshake interactions between people          \\ \midrule
\multirow{2}{*}{$\mathbb{Q}$} & $q_g$               & People belonging to the~same group \\ 
                   & $q_m$               & People wearing masks               \\ \bottomrule
\end{tabularx}

\label{tab:threatparams}
\end{table}

First, the~mask-wearing probability $q_m$ was used to quantify the~effect of masks in~transmission. Furthermore, people belonging to the~same group ($q_g$) have a~similar effect on transmission, since it~is assumed that the~disease spread between them does not increase depending on what is happening in~the~video frame (it is more likely they were both infected or not, even before coming into the~frame). The~values of $q_j$ are in~the~range $[0,1]$. $\epsilon_j \geq 1$ is used as~a~tuneable parameter that dictates the~influence of a~particular parameter $q_j$ on~the~overall threat level. A~higher $\epsilon_j$ value gives a~lower significance to the~corresponding $q_j$ in~calculating the~total threat $T(t)$. Because~the influence of various factors varies depending on variations and different pandemics, the~$\epsilon_j$ option can be used to change the~influence of various parameters, and~new parameters can be added to the~threat-level equation based on consultations with appropriate~authorities.

By substituting the~parameters and setting $\epsilon_m = 2.0$, $\epsilon_g = 1.0$, the~equation was rewritten as follows,
\begin{equation}
    T_{v_1,v_2}(t) = \left( p_h + p_d\right)\left(2.0 - q_m\right)\left(1.0 - q_g\right)
    \label{eqn:threat-our}
\end{equation}

When analyzing the~threat equation in~Equation~(\ref{eqn:threat-general}), it~can be noted that when the~secondary parameter probabilities decrease (i.e., $q_j$), the~effect of the~multiplicative term $(\epsilon_j - q_j)$ is higher. This~implies that the~effects of the~primary parameters $p_j$ to the~threat of the~given scene are compounded when the~two persons have worsening secondary parameters (i.e., are not wearing masks or when they are of different groups). It~can also be observed that \eqref{eqn:threat-our} does not carry any terms with the~$p_dp_h$ product. This~could be intuitively understood because shaking hands requires them to be physically close, and thus, incorporating this term is redundant. While \eqref{eqn:threat-our} is tuned for the~implemented system, the~generic form \eqref{eqn:threat-general} can incorporate any number of parameters being extracted from a~video~scene.

\section{Evaluation}

In this section, we~discuss the~methodology used to evaluate the~system.
The proposed solution was executed on a~chosen set of datasets as the~input, and the~results were evaluated using different metrics. 
The following subsections describe the~datasets, the~metrics, and~the evaluation execution process in~detail. 

\subsection{Datasets}

Existing public datasets such as MOT~\cite{mot-dataset-2020,mot-dataset-2016,mot-dataset-2015} and UT-interaction~\cite{UT-Interaction-Data} were chosen to evaluate the~performance of the~individual components of the~system. However, there are no existing datasets to perform a~holistic analysis. Thus, in~order to analyze this, a~new dataset was created from the~University of Peradeniya premises, which is referred to as the~UOP~dataset. \par

The \textbf{{multiple object tracking (MOT)}} datasets are a~set of image sequences with annotations for people localization and people IDs. Three datasets~\cite{mot-dataset-2020,mot-dataset-2016,mot-dataset-2015} were used to evaluate the~capability of an~algorithm to uniquely identify and track a~person through a~video.

\textls[-15]{The \textbf{{University of Texas--Interaction (UTI)}} \cite{UT-Interaction-Data, UT2} dataset comprises twenty video sequences of human interactions in~two or four-people settings. The~actions in~the~dataset include handshake, punch, point, kick, push and hug where each video spans roughly 1~min.}

The \textbf{{UOP dataset}} \cite{HandsOff} is a~collection of ten video sequences that were collected from the~University of Peradeniya premises by enacting a~scene with human interactions such as handshakes, close contacts, and~grouped conversations. These~videos were recorded by a~wall-mounted CCTV camera in~the~university corridor and waiting area. The~video consists of either four or five persons, with~each video spanning 1 min. The~ground truth for this dataset was annotated manually for training and~evaluation.

\subsection{Evaluation~Metrics}

The outputs were evaluated on~the~given datasets based on~the~metrics average precision (AP) and the~mean average precision (mAP). mAP is the~key metric used in~evaluating detector performance in~prominent object detection tasks such as the~PASCAL VOC challenge~\cite{pascalvoc}, COCO detection challenge~\cite{COCO} and the~Google Open Images competition~\cite{openimages}.

The average precision (AP) is the~precision value averaged across different recall values between 0 and 1~\cite{map-paper}. This~was computed as the~area under the~curve (AUC) of the~precision vs. recall curve, which was plotted as~a~function of the~confidence threshold of detection with a~fixed intersection over union (IoU) for the~bounding box threshold~\cite{padilla2}.

\subsection{Model~Evaluation}

\subsubsection{People~Detection}
The people detection component used here is the~YOLO network, which is a~well-established detector. Hence, no modifications were introduced to this segment of the~detector. The~YOLOv4 model which was used here is extensively compared in~terms of frame rate and mAP in~\cite{shinde2018yolo}. \par

\subsubsection{Group~Identification}
The group identification component was evaluated using the~existing MOT datasets. Since the~ground truth for the~datasets considered in~this work do not contain~the~group annotated information, an~alternative methodology was required for evaluation. For~this purpose, a~visual inspection of frames was used to determine if~two individuals belonged to the~same group in~a~given~frame. \par

\subsubsection{Mask~Detection}
The mask detection component requires localized information about masks. Thus, the~UT-interaction dataset was re-annotated. However, this dataset only consists of unmasked faces, and~as such, the~annotated UOP dataset was used together with the~UT-interaction dataset to train and evaluate the~mask detection component. The~17 videos from the~UT-interaction dataset and the~five videos from the~UOP dataset were used for training. The~dataset was annotated with the~two class information: namely, masked and unmasked faces in~frames, where the~faces were visible and the~presence of masks can be interpreted by a~human.

The mask detection model was evaluated using both the~AP and mAP measures. First, the~model's ability to localize the~faces was determined by measuring the~AP of the~localization component of the~models disregarding the~class~labels.

Next, the~performance of the~model in~terms of both the~localization and the~accuracy was determined by the~mean average precision (mAP) value. Note that since both the~classes correspond to the~same object (i.e., faces), this two-metric evaluation process helps us identify the~specific shortcomings of the~model considered. For~instance, a~high AP and a~low mAP show poor mask detection (classification), whereas~a~high accuracy and low mAP denote poor face~localization.

\subsubsection{Threat Level Assessment (End-to-End System)}

The threat level quantification algorithm was tested on~the~three datasets mentioned earlier. Since there is no publicly available ground truth for videos for this parameter, the~results of the~algorithm were evaluated by comparison with expert human input. For~this purpose, 462 samples of frame pairs from video sequences were chosen. The~system was then evaluated by observing the~increment/decrement of the~inferred threat level $T(t)$ and comparing the~results with the~expert human input. The~performance of the~full system is evaluated using accuracy, precision, and~recall. \par

\begin{figure}[H]
    \includegraphics[width=.99\textwidth]{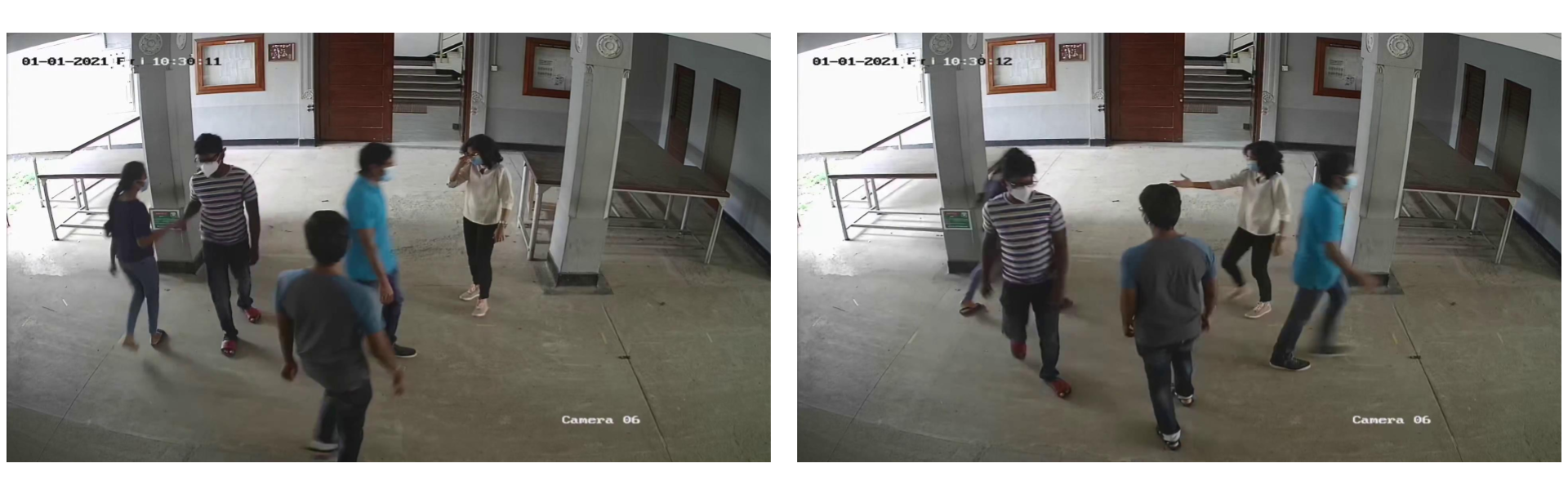}
    \caption{Example frames that were removed from full system evaluation due to disparity in~human expert~responses.}
    \label{fig:removed-system-eval}
\end{figure}

The expert responses were obtained by showing a~pair of frames and asking if~the~threat of \covid\ spread has increased or decreased from the~first frame to the~second. Since a~high disparity in~identifying the~impact of \covid\ spread can exist amongst human experts in~certain instances, ground truth cannot be established for such pairs of frames. To~identify such instances, a~thresholding minimum majority required to establish ground truth was set as 70\%, and all frame pairs with a~higher disparity (i.e., less than 70\% majority) for any given choice were removed. In~the evaluation conducted, five such frame pairs were identified and removed. One such frame pair is shown in~Figure~\ref{fig:removed-system-eval} to conclude this factor. As~it can be observed, it~is difficult to assess the~change in~threat for \covid\ spread across these two~frames.\par

\vspace{-6pt}

\section{Results and~Discussion}
The proposed system was implemented using the~Python programming language alongside and~Tensorflow and OpenCV libraries. The~system is deployed on a~high-performance computing server with NVIDIA GPU. The~output of each component of the~system as well as the~final output of the~entire system are discussed~below.

\subsection{People Detection and~Tracking}

\begin{figure}[H]
    \includegraphics[width=0.95\textwidth]{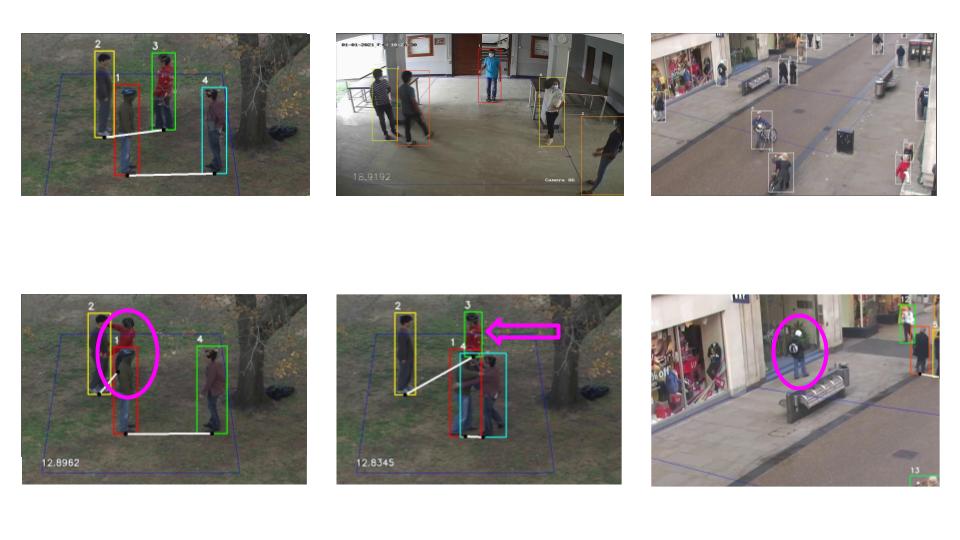}
\vspace{-10pt}
    \caption{{{Results of people }detection. (\textbf{Top row})---cases where the~people detection model is successful. (\textbf{Bottom row})---instances where people detection is erroneous. Undetected people are marked by the~purple oval. The~green bounding box (marked by the purple arrow) does not span the~full~person.}}
    \label{fig:res-yolo}
\end{figure}

The results shown in~Figure~\ref{fig:res-yolo} are indicative of the~performance of the~human detection and tracking segment of the~proposed system. The~first row shows a~sequence of frames where people are detected properly and tracked with unique IDs. However, the~model fails to perform satisfactorily in~specific scenarios. The~bottom row gives examples of \textls[-15]{the~cases in~which the~model can fail. From~left, (1) a~person is not being identified because of occlusion}, (2) the~identified bounding box is smaller than the~person due to occlusion, and~(3) a~person is going undetected due to the~lack of contrast with the~background. The~model has an \mbox{mAP = 65\%.\par}

\vspace{-10pt}

As observed in~Figure~\ref{fig:res-yolo}, a~given frame from the~output consists of multiple markings. The~blue quadrilateral on~the~ground is the~reference  used for perspective transformation. The~people detected are identified by uniquely colored rectangular bounding boxes. The~location of each person in~the~2D plane is marked using a~black dot on~the~bottom edge of the~respective bounding box. The~threat level for the~given frame is numerically displayed. Further details of the~relevant markings will be discussed in~the~subsequent~sections. \par

\subsection{Distance~Estimation}

\begin{figure}[H]
    \includegraphics[width=0.97\textwidth]{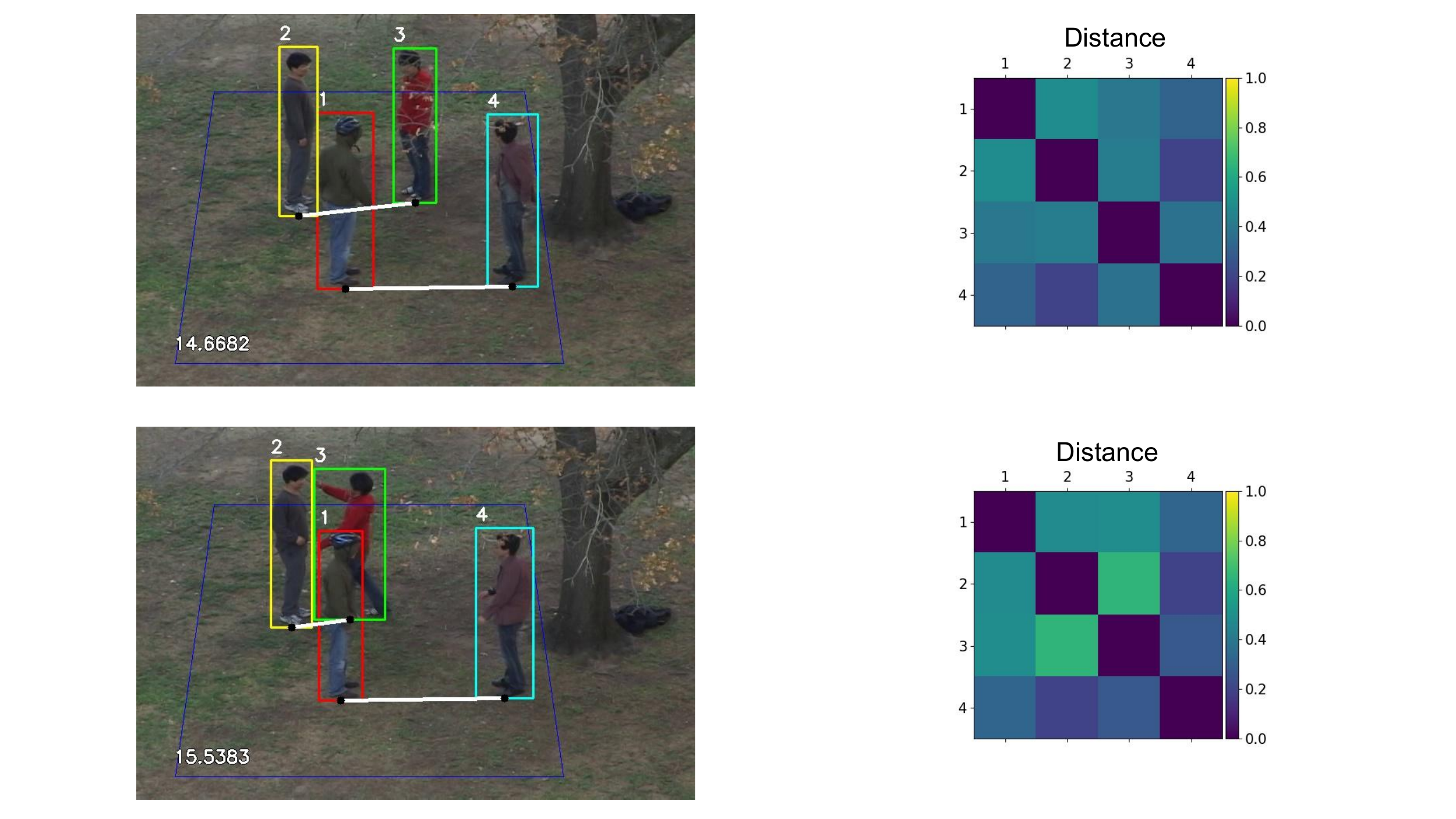}
    \caption{Distance estimation~results.}
    \label{fig:res-distance}
\end{figure}

A scene consisting of four people from the UTI dataset is considered in~Figure~\ref{fig:res-distance} to show how the~distance between people contributes to the~threat level. The~distance between people is given by the~distance activity matrix shown beside each frame in~Figure~\ref{fig:res-distance}. Each element (square) in~the~activity matrix denotes the~proximity between the~person IDs corresponding to the~row and column indices. The~color changes to a~warmer shade (yellow) when the~people are closer, and~it becomes a~colder shade (blue) when they are farther~away.

Considering the~frames in~Figure~\ref{fig:res-distance}, the~person ID 2 and 3 can be observed to be closer in~the~second frame than in~the~first frame. This~causes a~higher contribution to the~threat level between them in~the~second frame and a~lower contribution to the~threat level in~the~first frame. This~is seen in~the~distance activity matrix by the~blue shade turning to cyan, indicating closer proximity between those persons. The~reader's attention is drawn to the~threat level shown in~each frame. As~it can be observed, when the~distance activity matrix lightens up, the~threat level has also~risen. \par

\textls[-15]{The errors in~people detection (Figure~\ref{fig:res-yolo}) can propagate to the~distance estimation. While people going fully undetected is usually handled by interpolation, predicted bounding boxes becoming cropped due to the occlusion of feet leads to a~faulty prediction for where the~person's feet are. Therefore, the~calculation of distance between people becomes~erroneous.}

\subsection{Group~Identification}

\begin{figure}[H]
  \includegraphics[width=.99\textwidth]{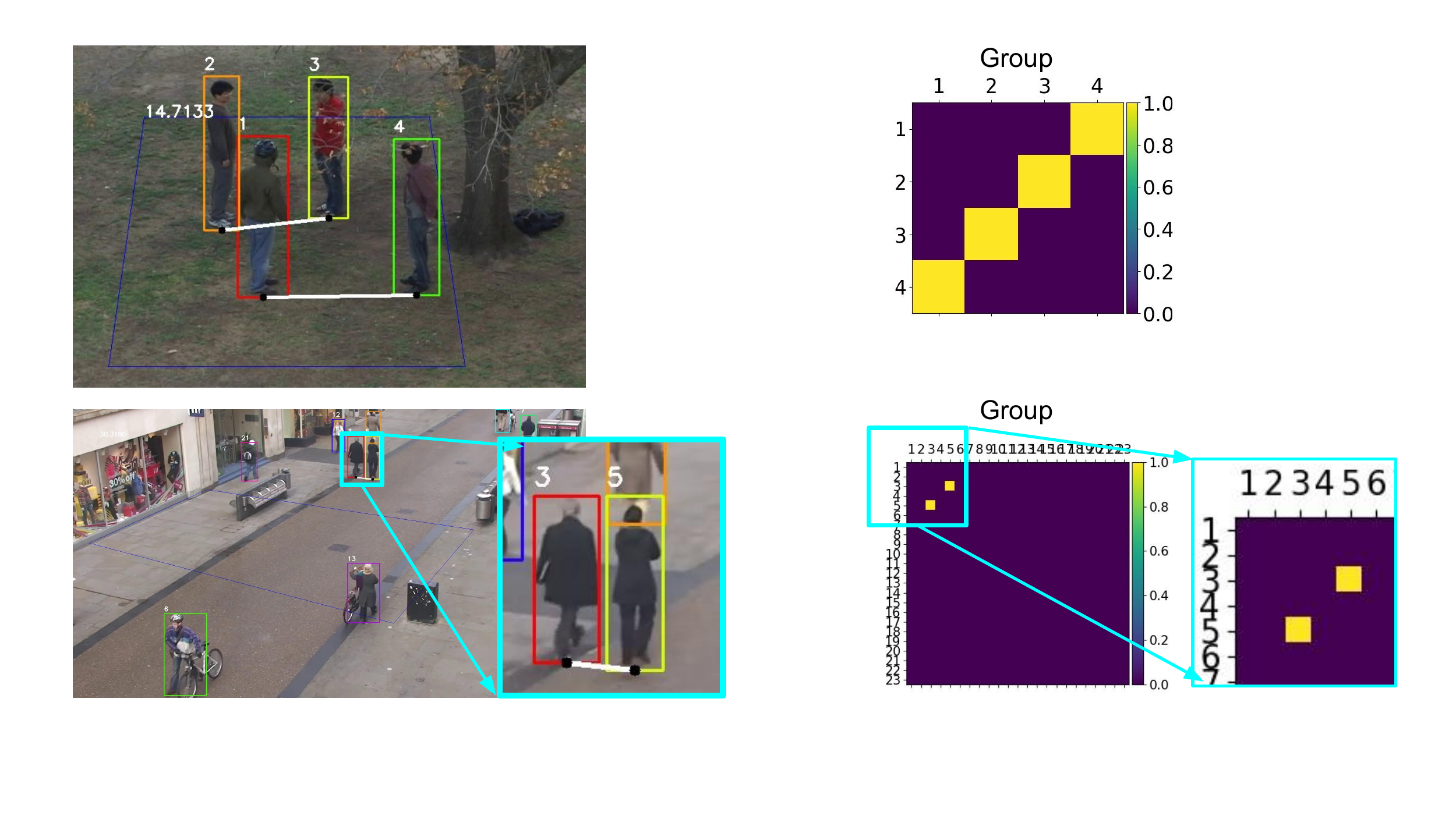}
    \caption{{Group identification}. (\textbf{Left}): video frames where groups of people are denoted by white lines connecting individuals. (\textbf{Right}): group activity matrices showing people belonging to the~same group by yellow and else~blue.}
    \label{fig:res-group}
\end{figure}

\textls[-20]{The results for a~few frames for the~group identification model are shown in~Figure~\ref{fig:res-group}}. An~example from the~UTI dataset and Oxford towncenter datasets is shown here. The~frames with the~persons detected are shown on~the~left and the~group activity matrices showing the~group characteristics are shown on~the~right. If~two people are of the~same group, the~group activity matrix element corresponding to the~row and column of the~IDs of these two persons is shown in~yellow, and otherwise, it~is shown in~blue. The~people of the~same group are also joined by a~white line in~the~original frame to show~this.

\subsection{Mask~Detection}

\begin{figure}[H]
\begin{subfigure}{.33\textwidth}
    \centering
    \includegraphics[width=\textwidth]{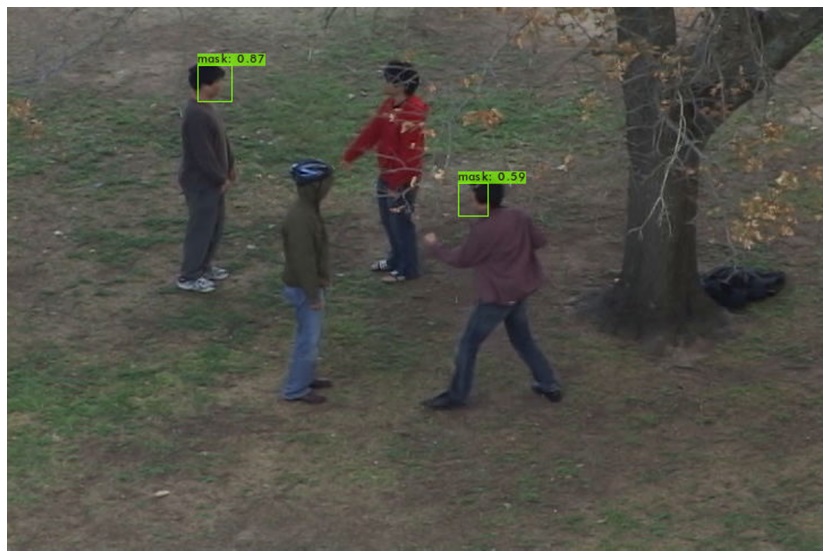}
    \caption{\centering}
\end{subfigure}
\begin{subfigure}{0.33\textwidth}
    \centering
    \includegraphics[width=\textwidth, height= 3.1cm]{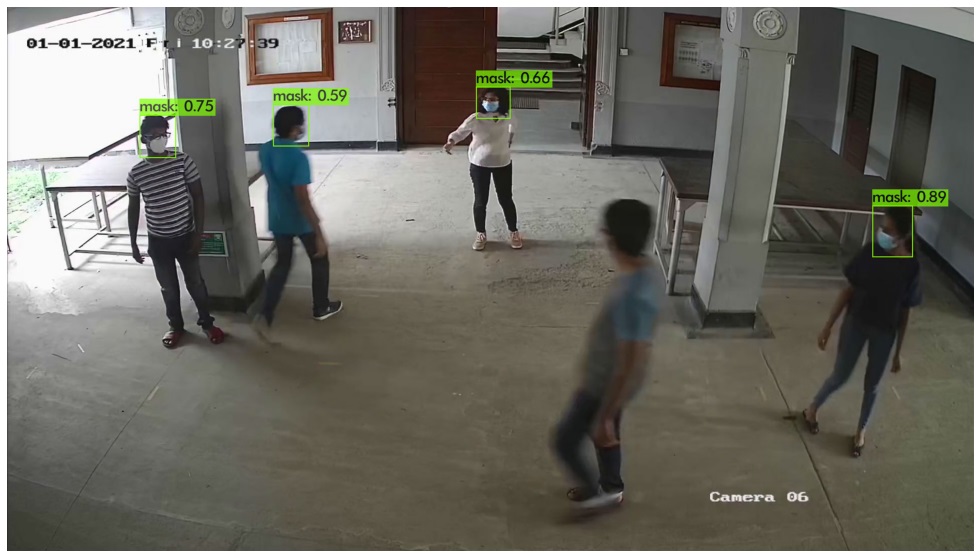}
    \caption{\centering}
\end{subfigure}%
\begin{subfigure}{.33\textwidth}
    \centering
    \includegraphics[width=\textwidth]{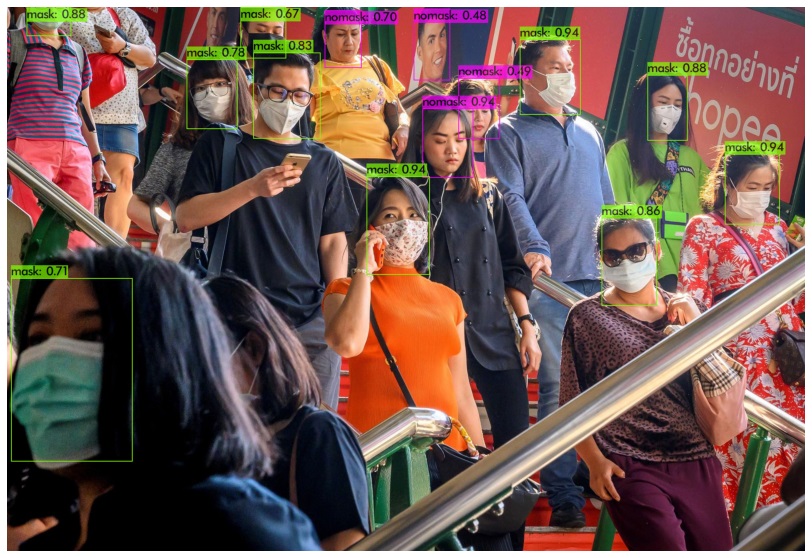}
    \caption{\centering}
\end{subfigure}
    \caption{Mask detection detection~examples. {(\textbf{a}) UTI~dataset; (\textbf{b}) UOP~dataset; (\textbf{c}) Moxa3K~dataset.}}
    \label{fig:res-mask}
\end{figure}

Figure~\ref{fig:res-mask} shows the~performance of the~system in~detecting the~presence/absence of masks. One example from the~UTI, UOP, and~Moxa3K datasets is shown. Overall the~system performs well while dealing with high-resolution images (Moxa3K). However, as~the resolution drops (UTI/UOP), the~efficacy reduces drastically. This~can be observed in~\tref{tab:res-mask}, which lists the~numerical evaluation metrics (AP and mAP) for localization on different datasets. Another prominent failure case is when the~proposed system is unable to detect the~presence or the~absence of masks when people face away from~cameras.

\unskip

\unskip

\begin{table}[H]
    \caption{Performance metrics of the~mask~detection.}
   \newcolumntype{C}{>{\centering\arraybackslash}X}
\begin{tabularx}{\textwidth}{CC}
\toprule
        \textbf{Dataset} & \textbf{AP/mAP /\%} \\
        \midrule
        UT-interaction (Unmasked) & 29.30 \\
        UOP (Masked) & 41.47 \\
        Moxa3K & 81.04 \\
        \bottomrule
    \end{tabularx}
    \label{tab:res-mask}
\end{table}
\unskip

\subsection{Threat Level Assessment (End-to-End System)}

\begin{figure}[H]
    \includegraphics[width=0.90\linewidth]{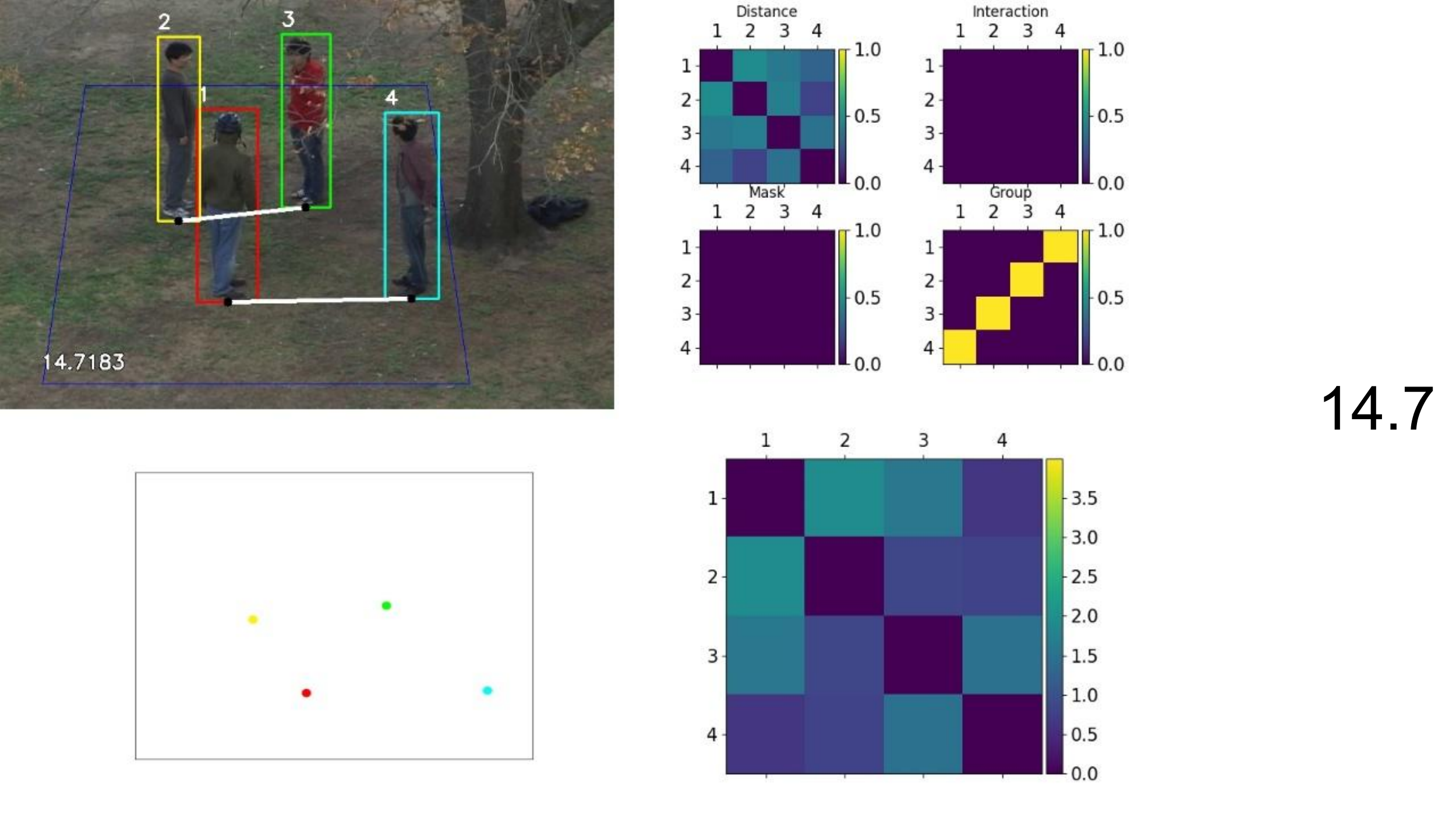}
    \caption{{Full system} result of UTI interaction dataset at $t_1$.}
    \label{fig:res-full-uti-1}
\end{figure}

\begin{figure}[H]
    \includegraphics[width=0.90\linewidth]{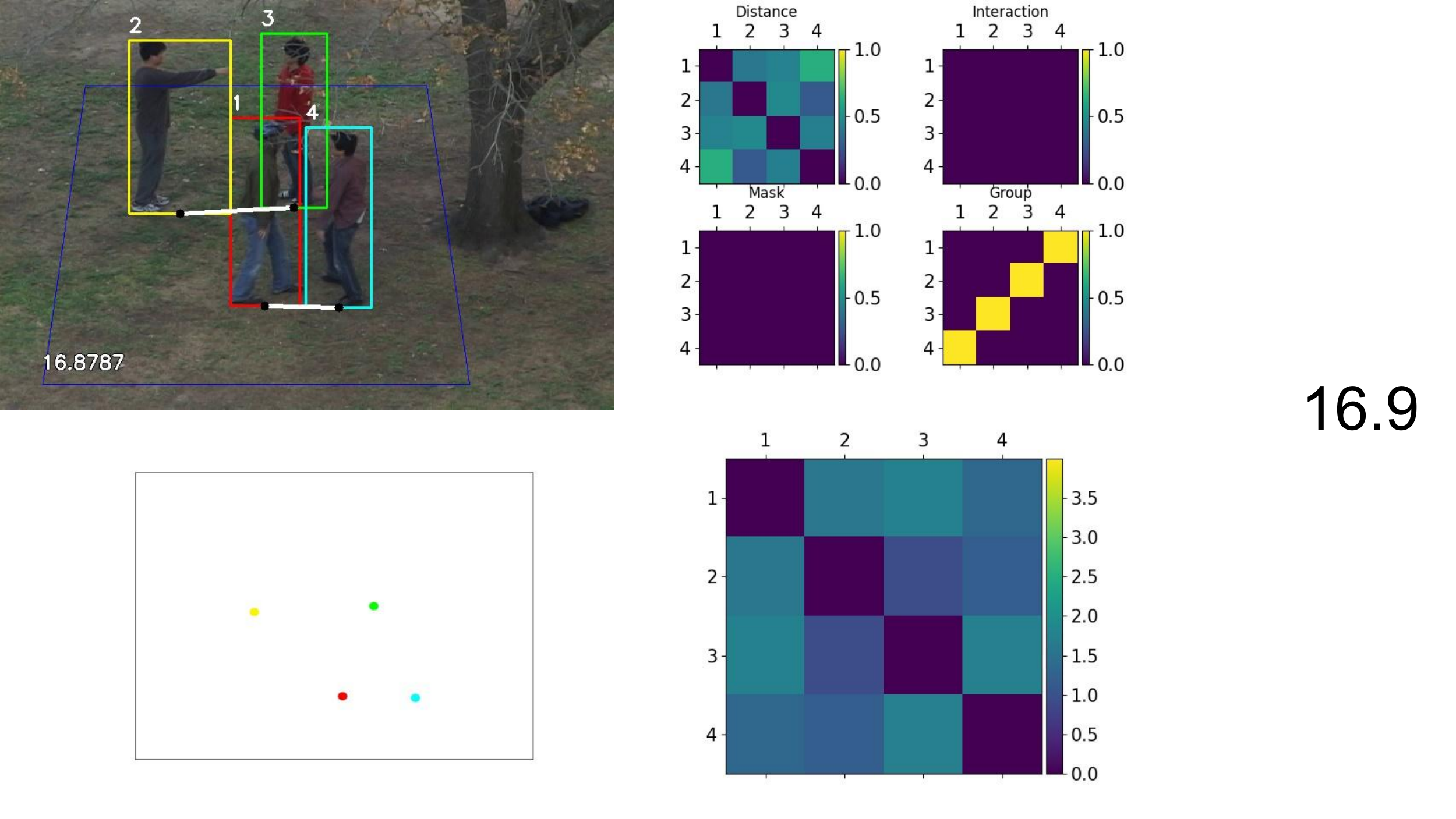}
    \caption{{Full system} result of UTI interaction dataset at $t_2$.}
    \label{fig:res-full-uti-2}
\end{figure}

\begin{figure}[H]
    \includegraphics[width=0.90\linewidth]{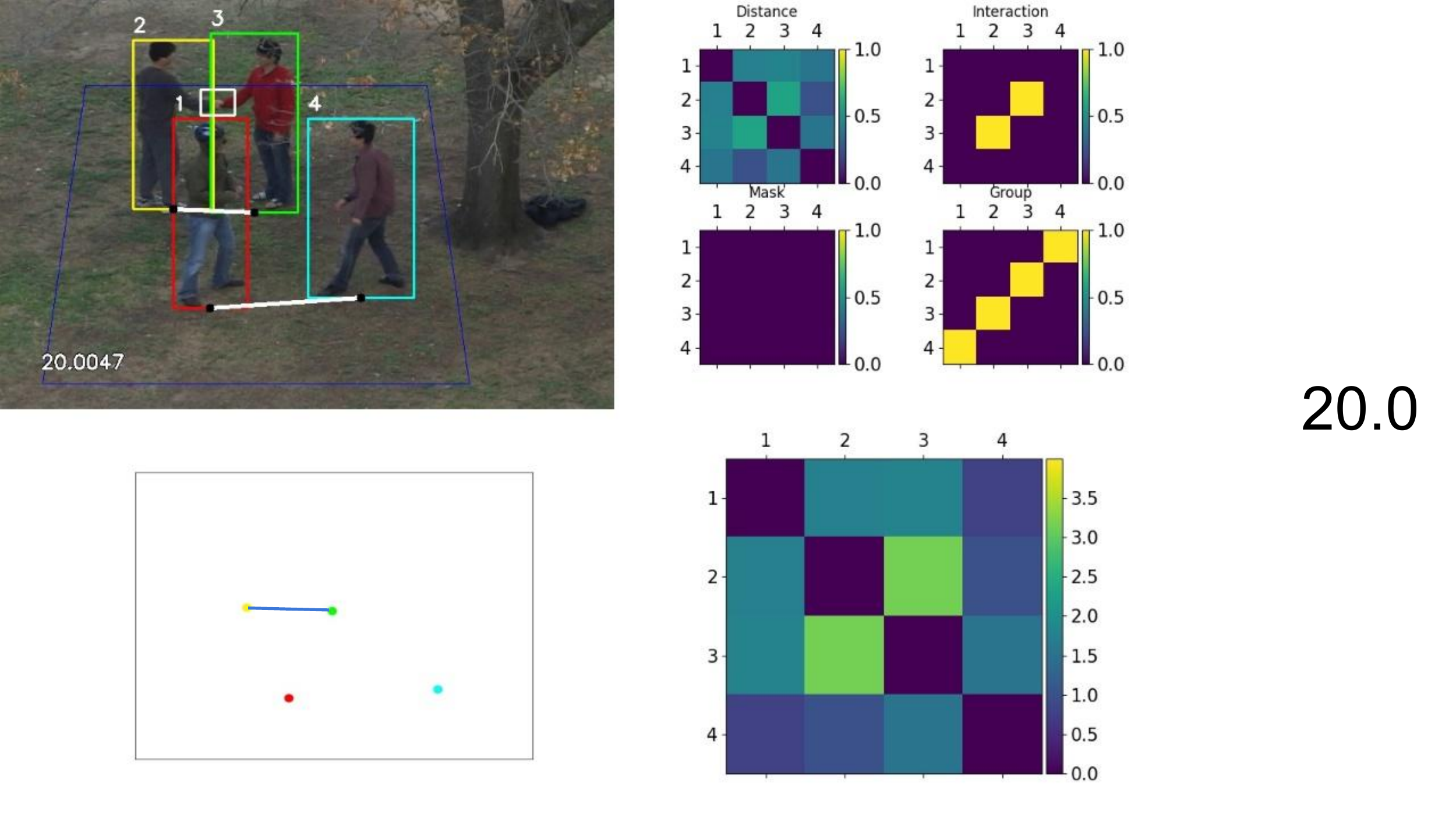}
    \caption{{Full system} result of UTI interaction dataset at $t_3$.}
    \label{fig:res-full-uti-3}
\end{figure}

\begin{figure}[H]
    \includegraphics[width=1\linewidth]{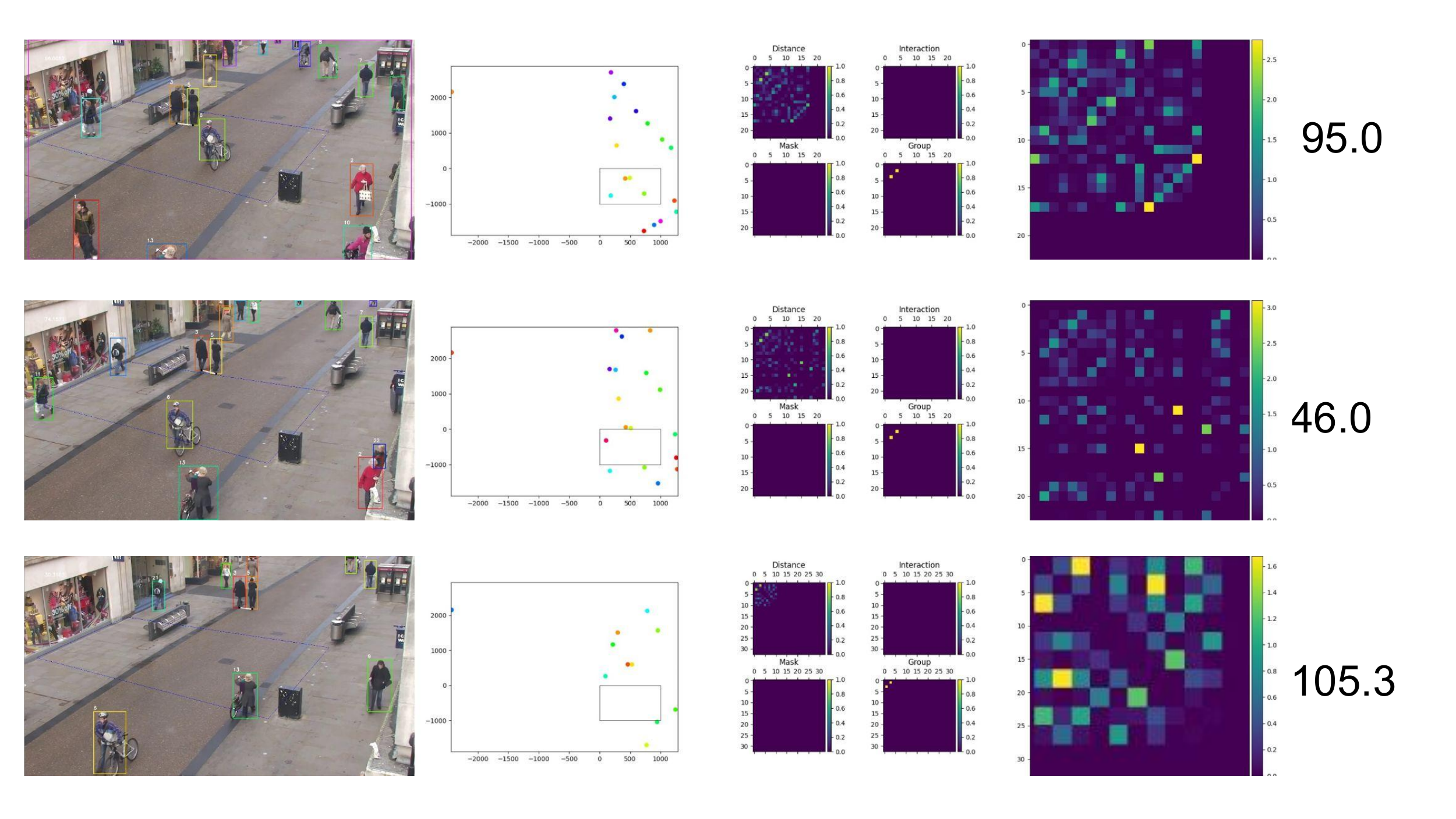}
    \caption{{Full system} result of oxford~dataset.}
    \label{fig:res-full-oxford}
\end{figure}

To evaluate the~proposed system performance, the~threat level metric provided for each frame of a~given scene is evaluated across multiple frames. The~successful output of this value is evaluated by the~full system for both datasets UTI (Figures \ref{fig:res-full-uti-1}--\ref{fig:res-full-uti-3}) and~Oxford (Figure~\ref{fig:res-full-oxford}). It~should be noted that it~is not the~absolute values of the~threat level that are significant but the~increments or decrements between the~frames.

Considering Figures~\ref{fig:res-full-uti-1}--\ref{fig:res-full-uti-3}, it~can be observed that the~threat level increases from top to bottom frames as 14.7, 16.9 and 20.0. From~the first frame to the~second frame (Figure~\ref{fig:res-full-uti-1} to Figure~\ref{fig:res-full-uti-2}), we~can see the~distance activity matrix brightening in~the~right top and left bottom edges. This~is due to the close proximity of persons ID 1 and 4. This~leads to an~increase in~the~threat level of the~frame by 16.9 $-$ 14.7 = 2.2. Similarly, when looking at the~first and third frames (Figure~\ref{fig:res-full-uti-1} to Figure~\ref{fig:res-full-uti-3}), this time, the~interaction activity matrix brightens up in~the~third frame due to the~handshake interaction in~this frame. This~also leads to an~increase in~threat level, which is by 20.0 $-$ 14.7 = 5.3. It~is also clearly observed in~the~threat activity matrix for the~third frame in~Figure~\ref{fig:res-full-uti-3}, where the~center squares brighten up to show a~significant threat between persons 2 and 3. This~increment (of 5.3) in~the~threat level is higher than the~previous comparison (of 2.2) in~Figure~\ref{fig:res-full-uti-1}, and~Figure~\ref{fig:res-full-uti-2} since the~handshake interaction poses a~higher threat than proximity alone. The~same can be observed by comparing the~second and third~frames. \par

A simpler situation is analyzed in~Figure~\ref{fig:res-full-oxford}. Here, there are only two people belonging to the~same group, and they are present in~the~video throughout the~time. However, there are no physical interactions such as shaking hands. Therefore, the~only parameter that dictates the~threat level is the~number of people and their interpersonal distances in~each frame. When analyzing Figure~\ref{fig:res-full-oxford}, the~people in~the~first frame are moving away from each other until the~second frame. This~is why the~threat level goes down from 95.0 to 46.0 from the~first frame to the~second. In~the third frame, new people come into the~frame, and they move closer to each other. Therefore, an~increase in~the~threat level of 105.3 is observed. However, this dataset does not contain~a~rich set of scenes to evaluate all components of the~proposed~system.

\unskip

\unskip

\unskip

\subsection{Full System~Evaluation}

\textls[-15]{The performance of the~full system in~comparison to human expert responses is provided in~\tref{tab:system-perf} in~terms of accuracy, precision, and~recall. The~complete system is evaluated on both the~UTI and UOP datasets. However, it~should be noted that the~UTI dataset does not contain anyone wearing masks, and~hence, the~mask detection component does not contribute to the~threat calculation here. It~can be noted that the~system performance is not biased toward either dataset and is able to generalize with considerable accuracy of nearly 76\%.} \par

\begin{table}[H]
    \caption{Full system~performance.}
    \newcolumntype{C}{>{\centering\arraybackslash}X}
\begin{tabularx}{\textwidth}{CCCC}
\toprule
         \textbf{Test} & \textbf{Accuracy} & \textbf{Precision} & \textbf{Recall}\\
         \midrule
         UTI dataset & 75\% & 75\% & 75\% \\
         UOP dataset & 76\% & 85\% & 79\%\\
         Overall & 76\% & 81\% & 77\%\\
         \bottomrule
    \end{tabularx}
    \label{tab:system-perf}
\end{table}

\begin{figure}[H]
    \includegraphics[width=0.97\textwidth,height=4.8cm]{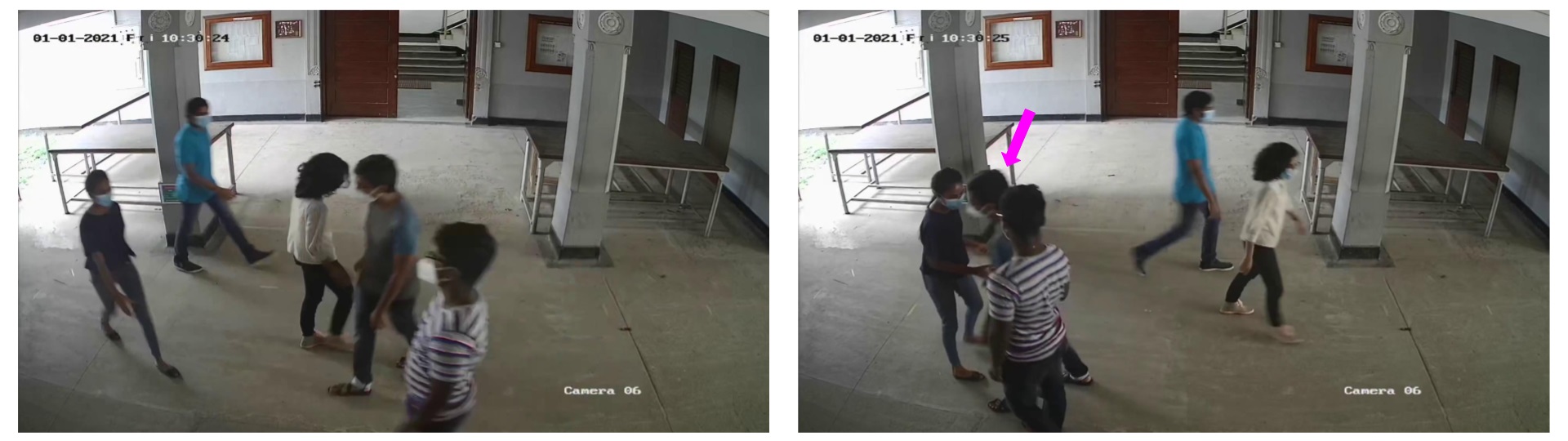}
    \caption{{Failure case} 1 threat level interpretations. System output for threat---Decreases, Human expert opinion on threat---Increases.}
    \label{fig:sys-fail1}
\end{figure}

\begin{figure}[H]
    \includegraphics[width=0.97\textwidth,height=4.8cm]{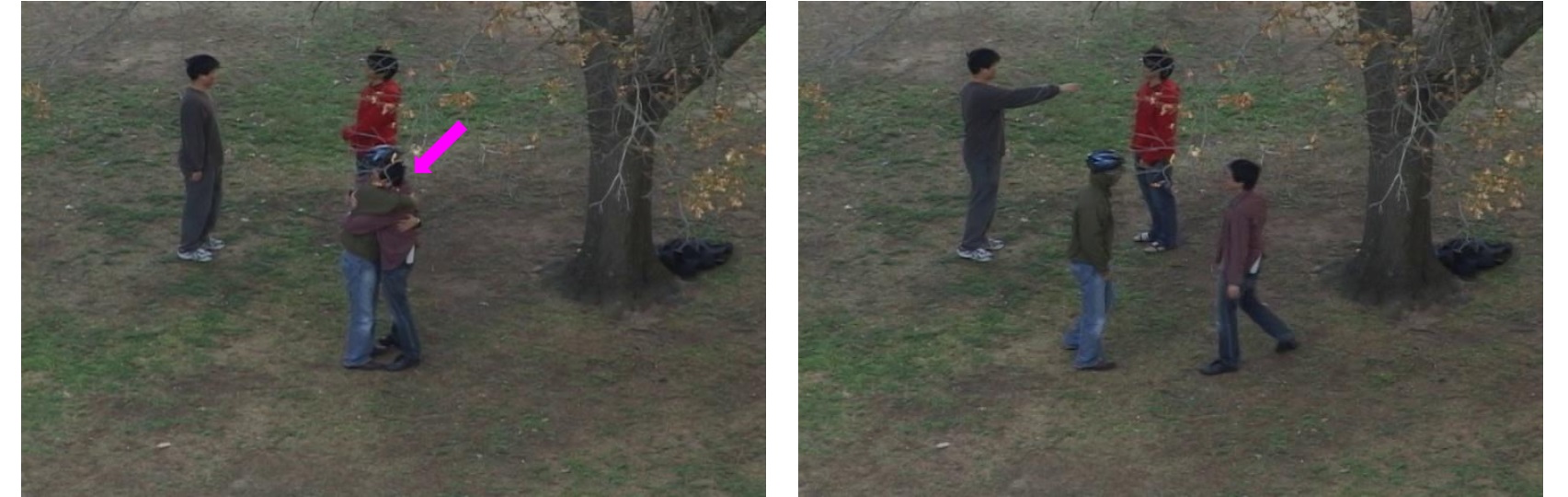}
    \caption{{Failure case} 2 threat level interpretations. System output for threat---Increases, Human expert opinion on threat---Decreases.}
    \label{fig:sys-fail3}
\end{figure}

\begin{figure}[H]
    \includegraphics[width=0.97\textwidth,height=4.8cm]{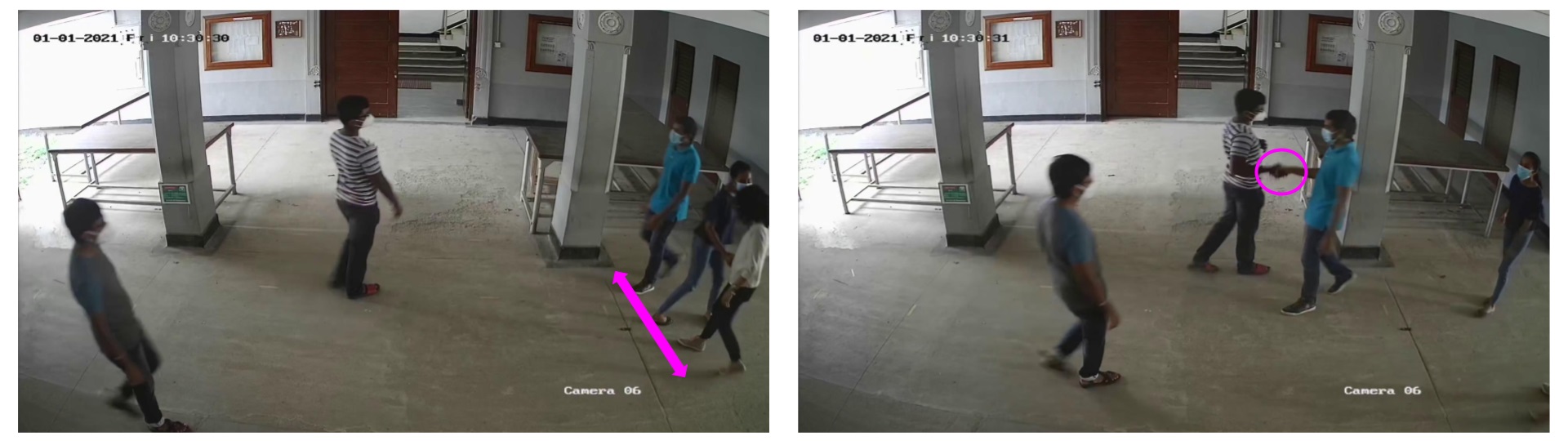}
    \caption{{Failure case} 3 threat level interpretations. System threat evaluation output---Decreases, Human expert opinion output---Increases.}
    \label{fig:sys-fail2}
\end{figure}

\begin{figure}[H]
    \includegraphics[width=0.97\textwidth,height=4.8cm]{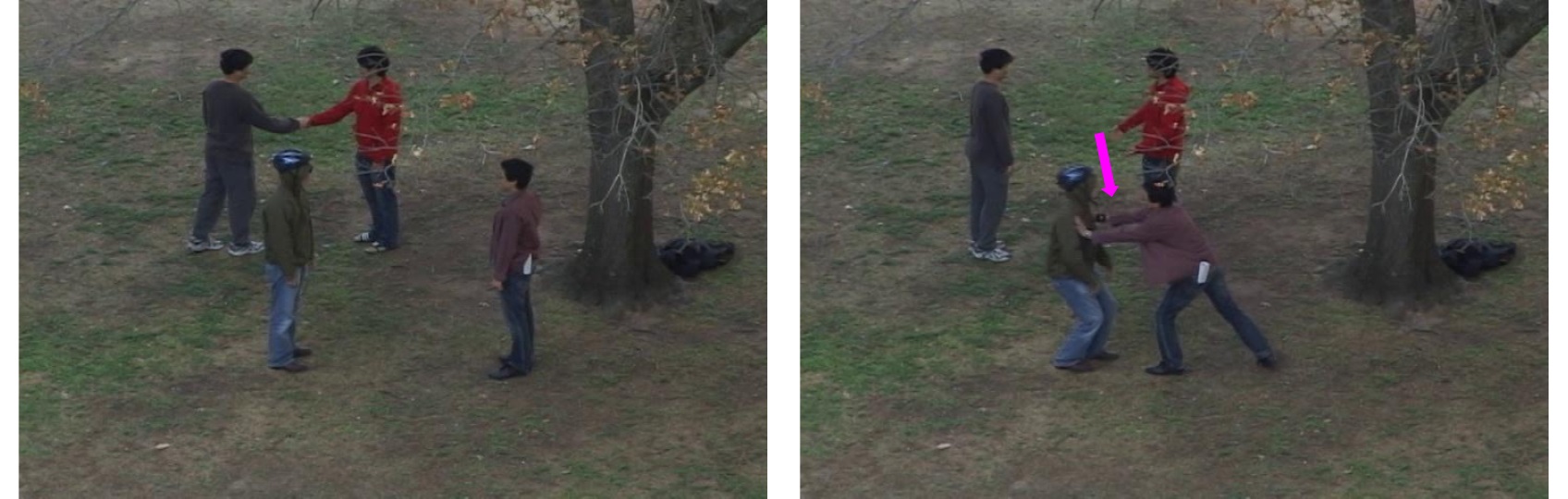}
    \caption{{Failure case} 4 threat level interpretations. System threat evaluation output---Decreases, Human expert opinion output---Increases.}
    \label{fig:sys-fail4}
\end{figure}

A few of the~notable failure cases of the~system are shown in~Figures~\ref{fig:sys-fail1}--\ref{fig:sys-fail4}, where the~threat level predicted was contrary to human expert opinion. Out of the~four cases shown here, three of them failed to evaluate the~proper threat value due to a~failure in~one of the~components in~the~system pipeline. In~Figure~\ref{fig:sys-fail1}, the~person indicated by the~purple arrow was not detected by the~person detection model due to occlusion. Similarly, in~Figure~\ref{fig:sys-fail3}, the~two individuals hugging are detected as~a~single person. Since it~is the~proximity of the~three individuals in~Figure~\ref{fig:sys-fail1} and the~hugging individuals in~Figure~\ref{fig:sys-fail3} that pose a~high threat to \covid\ spread, the~system fails to reflect this, deviating from the~expert opinion. In~Figure~\ref{fig:sys-fail2}, the~high proximity of the~individuals in~the~first frame results in a~high threat value for the~first frame. However, the~handshake interaction model fails to detect the~interaction in~the~second frame, hence leading to a~lower threat level output by the~system and hence failing to identify the~increase in~threat for \covid\ spread. In~the case of Figure~\ref{fig:sys-fail4}, since the~system design was not accounted for incidents such as~a~pushing action as in~the~second frame, the~system provides a~higher threat value for the~first frame contrary to human expert~opinion. \par

\unskip

\unskip

\unskip

\begin{figure}[H]
    \includegraphics[width=0.97\textwidth,height=4.8cm]{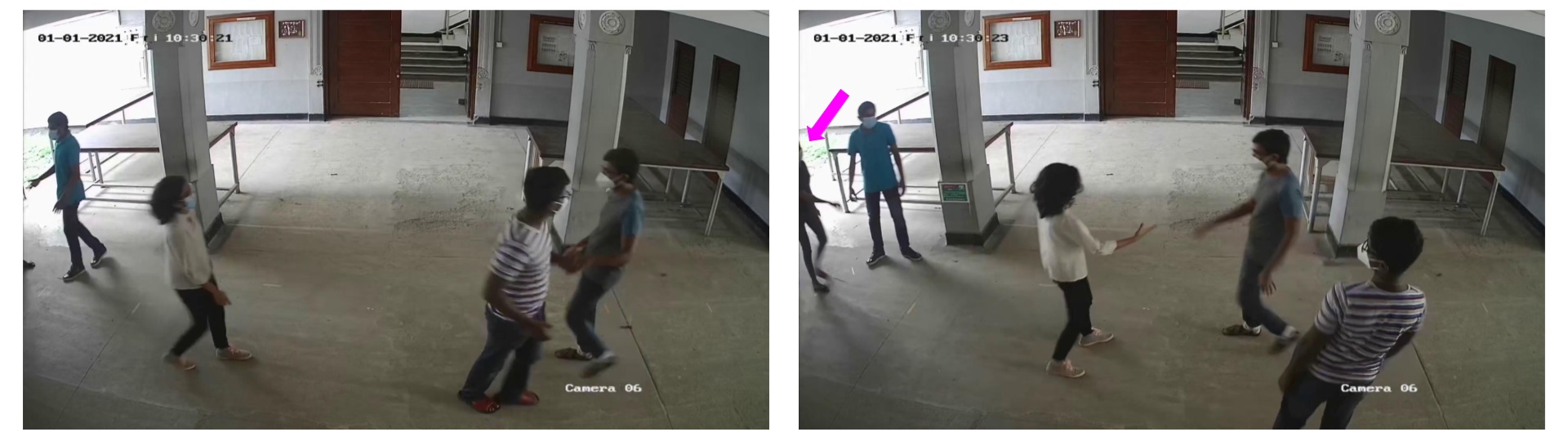}
    \caption{{Edge case} 1 threat level interpretations. System threat evaluation output---Increases, Human expert opinion output---Decreases.}
    \label{fig:sys-plus}
\end{figure}

\begin{figure}[H]
    \includegraphics[width=0.97\textwidth,height=4.8cm]{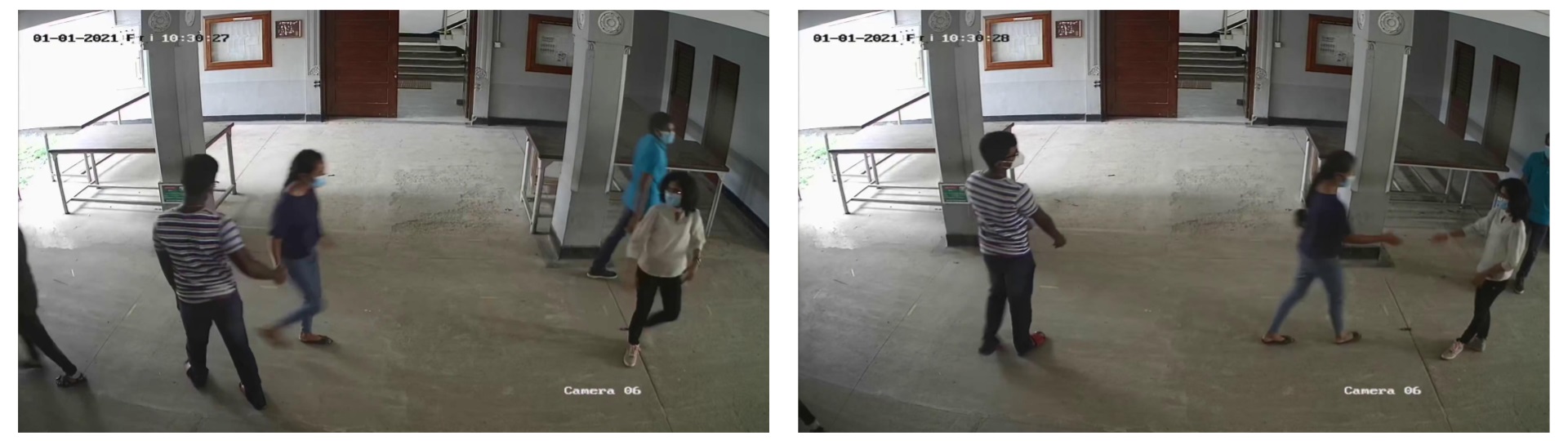}
    \caption{{Edge case }2 threat level interpretations. System threat evaluation output---Increases, Human expert opinion output---Decreases.}
    \label{fig:sys-unexpl}
\end{figure}

\textls[-15]{However, there were a few rare cases where in~retrospect, the~system output was more plausible or instances where the~failure of the~system was unexplained. Considering Figure~\ref{fig:sys-plus}, the~ground truth from human expert opinion was that the~threat level decreases, which is explained by the~handshake interaction in~the~first frame, which is a~serious violation of social distancing protocols. However, the~system output for threat value increases significantly in~the~second frame as~a~new person is identified in~the~far left. Since an~increase in~the~number of people and the closer proximity of this new person in~a~given space should also be accounted for, this leads to the~increased threat value predicted by the~system. Meanwhile, Figure~\ref{fig:sys-unexpl} is an~instance where the~system output states the~threat of \covid\ spread has increased, whereas human expert opinion is on~the~contrary. This~deviation by the~system is an~edge case where the~deviation is~unexplained.} \par

\unskip

\section{Conclusions}

An end-to-end solution utilizing CCTV footage to provide a~practical and versatile mechanism for monitoring crowds to identify possible instances of spreading COVID-19 is proposed in~this paper. The~proposed system detects people, their locations, their individual actions (wearing masks), and~their dyadic interactions (handshakes) using deep learning-based video interpretation. This~information is stored in~temporal graphs to enable further insights such as identifying social groups. Finally, these temporal graphs undergo a~more holistic interpretation using rule-based algorithms. This~analysis uncovers the~individual's contributions to the~spread of COVID-19 (not wearing masks) and~pairwise contributions (handshakes, staying close to others) on a~per frame basis. These~results are brought together to calculate the~total threat levels in~frames. Finally, these outputs are examined against expert human opinion. Consistent accuracies over 75\% across all datasets could be considered a~strong indication of the~robust performance of the~proposed~system.\par

Furthermore, this unified framework allows for the future incorporation of possible other future measures for curtailing the~spread of COVID-19 or any other epidemics impacting the~health and safety of society. Therefore, this proposed framework may be strengthened by incorporating additional COVID-19 specific features, and it~could be adapted and adopted for similar other scenarios as well that may benefit from video or CCTV-based non-intrusive~observations. \par 

The proposed system evaluation is limited by the~availability of datasets. While it~was tested  on existing datasets and a~newly collected dataset, testing this on a~wider range of diverse scenes is required. Moreover, threat level estimation is performed at the~frame level. As~a result, the~contribution of the~time of interactions such as approaching close to each other and shaking hands is not taken into account in~our system. Future work could improve the~system by processing the~time series of the~threat level (individual contributors such as interactions as well as the~total score) by techniques such as moving averages and~filtering. \par

Due to the~widespread use of CCTV cameras, the~proposed system is applicable to a~wide variety of real-world scenarios. The~decreasing performance-to-cost ratio of computing hardware 
has enabled even small organizations to acquire the~system. The~release of the~codebase as free and open source software (FOSS) can accelerate both third-party deployment and solution improvement. However, concerns about privacy, bias, and~fairness in~conducting analytics on people's CCTV footage should be addressed on an~individual basis in~accordance with the~rules and regulations of individual organizations and~countries.

\vspace{6pt} 



\authorcontributions{{Conceptualization}, G.J., J.H., S.S., R.G., P.E., V.H. J.E. and S.D.; Data curation, G.J., J.H. and S.S.; Formal analysis, G.J., J.H. and S.S.; Funding acquisition, R.G., P.E., V.H. J.E and S.D.; Methodology, G.J., J.H. and S.S.; Project administration, R.G., P.E., V.H. and J.E.; Software, G.J., J.H. and S.S.; Supervision, R.G., P.E., V.H., J.E. and S.D.; Validation, J.B.S., H.W.; Visualization, G.J., J.H., S.S., J.B.S., H.W.; Writing---original draft, G.J., J.H. and S.S.; Writing---review \& editing, G.J., J.H., S.S., J.B.S., H.W., R.G., P.E., V.H. and J.E. All authors have read and agreed to the published version of the manuscript.} 

\funding{{This work is funded by (1) International Development Research Centre (IDRC), Canada through grant number 109586-001 and (2) Lewis Power, Singapore. The APC was funded by IDRC as well.}}

\institutionalreview{{Not applicable}.}

\informedconsent{{Informed consent was obtained from all subjects involved in the study for all stages including dataset creation and system evaluation}.}

\dataavailability{{The data collecetd in this study could be requested from J.E. via email (ekanayakej@eng.pdn.ac.lk). The requests will be handled in a case by case basis.}}

\acknowledgments{{ 
GPU computing resources were provided by NVIDIA to the~Embedded Systems and Computer Architecture Laboratory (ESCAL), University of Peradeniya through the~GPU donation program. This~research was made possible through the~University of Peradeniya research infrastructure funded by the~residents of Sri Lanka through their contribution to public education.} The~comments and suggestions from the~anonymous reviewers has been instrumental in~revising this~paper.}

\conflictsofinterest{{The authors declare no conflict of interest. The funders had no role in the design of the study; in the collection, analyses, or interpretation of data; in the writing of the manuscript, or in the decision to publish the~results.}}

\abbreviations{Abbreviations}{
{Notations} 
  and description.\\

\noindent 
\begin{tabular}{@{}lp{11cm}}
\textbf{Notation} & \textbf{Definition} \\
    $V_{in}(t)$ & Input video~feed\\
    $F_p$, $\bar{F}_p$ & People detection and~tracking\\
    $F_d$ & Distance~estimation\\
    $F_g$, $\bar{F}_g$ & Group identification and~tracking\\
    $F_h$ & Identifying and localizing physical interaction (handshakes) \\
    $F_m$, $\overline{F}_m$ & Mask detection and~tracking\\
    $J_{i}(t)$ & Output of model $F_i$ \\
    $S_i(t)$ & State~information \\
    $bb_{pk}(t)$, $bbi_{pk}(t)$ & Bounding box encompassing person $k$ at time $t$ and bounding box encompassing person $k$ at time $t$ which is being tracked with their unique~index\\
    $bb_{hk}(t)$, $bbi_{hk}(t)$ & Bounding box encompassing handshake interaction $k$ at time $t$  and bounding box encompassing handshake interaction $k$ at time $t$ which is being tracked with their unique~index.\\
    $bb_{mk}(t)$, $bbi_{mk}(t)$ & Bounding box encompassing the~face of person $k$ at time $t$  and bounding box encompassing face of person $k$ at time $t$ which is being tracked with their unique~index\\
    $u, v$ & The 2D coordinates of the~center of the~bounding~box\\
    $h, r$ & The height and aspect ratio of the~bounding~box\\
    $R, R'$ & The coordinates of the~reference points in~the~video frame and two-dimensional floor plane,~respectively\\
    $M_T$ & Transformation matrix for the~perspective transform from CCTV perspective to floor~plane\\
    $s_{(i,t)}$ & Standing location of person $i$ at time $t$ in~the~CCTV~perspective\\
    $floorLocation_{(i,t)}$ & Standing location of  person $i$ at time $t$ in~the~floor~plane\\
    $dist_{(i,j,t)}$ & Distance between a~pair of people $i$ and $j$ at time $t$\\
    $P_i$ & Person $i$ in~the~frame\\
    $G(t)$ & Graph at time $t$\\
    $V(t)$ & Vertices of graph $G$ at time $t$ given by $\{v_1(t), v_2(t), \dots, v_n(t)\}$, each vertex corresponding to person $P_i$ with the~vertex parameters~embedded\\
    $E(t)$ & Edges of graph $G$ at time $t$ given by $\{e_{1,1}(t), e_{1,2}(t), \dots, e_{i,j}(t), \dots, e_{n,n}(t)\}$, where $e_{i,j}$ is the~edge between person(vertex) $i$ and $j$\\
    $T(t)$ & Threat level of frame at time $t$\\
    $\mathbb{P} = \{p_d, p_h\}$ & Primary parameters---set of parameters that have a~direct attribute to \covid\ transmission\\
    $\mathbb{Q} = \{q_g, q_m \}$ & Secondary parameters---set of parameters that are relevant to \covid\ transmission when two individuals are in~close~proximity\\
    $\epsilon_j$ & Tuneable parameter dictating influence of parameter $q_j$ on overall threat~level.\\
\end{tabular}}

\begin{adjustwidth}{-\extralength}{0cm}

\reftitle{References}

\end{adjustwidth}


\begin{thebibliography}{999}

\bibitem[Zhao \em{et~al.}(2020)Zhao, Yao, Wang, Zheng, Gao, Ye, Guo, Zhao, and
  Gao]{covid-pneumonia}
Zhao, D.; Yao, F.; Wang, L.; Zheng, L.; Gao, Y.; Ye, J.; Guo, F.; Zhao, H.;
  Gao, R.
\newblock A comparative study on~the~clinical features of coronavirus 2019
  (COVID-19) pneumonia with other pneumonias.
\newblock {\em Clin. Infect. Dis.} {\bf 2020}, {\em 71},~756--761.

\bibitem[Long \em{et~al.}(2020)Long, Brady, Koyfman, and
  Gottlieb]{covid-cardiovascular}
Long, B.; Brady, W.J.; Koyfman, A.; Gottlieb, M.
\newblock Cardiovascular complications in~COVID-19.
\newblock {\em  Am. J. Emerg. Med.} {\bf 2020}, {\em
  38},~1504--1507.

\bibitem[Ellul \em{et~al.}(2020)Ellul, Benjamin, Singh, Lant, Michael, Easton,
  Kneen, Defres, Sejvar, and~Solomon]{covid-neuro}
Ellul, M.A.; Benjamin, L.; Singh, B.; Lant, S.; Michael, B.D.; Easton, A.;
  Kneen, R.; Defres, S.; Sejvar, J.; Solomon, T.
\newblock Neurological associations of COVID-19.
\newblock {\em  Lancet Neurol.} {\bf 2020}, {\em
  {19}},~{767--783}. 

\bibitem[Lopez~Bernal \em{et~al.}(2021)Lopez~Bernal, Andrews, Gower, Gallagher,
  Simmons, Thelwall, Stowe, Tessier, Groves, Dabrera, Myers, Campbell,
  Amirthalingam, Edmunds, Zambon, Brown, Hopkins, Chand, and
  Ramsay]{delta-vaccine-efficacy}
Lopez~Bernal, J.; Andrews, N.; Gower, C.; Gallagher, E.; Simmons, R.; Thelwall,
  S.; Stowe, J.; Tessier, E.; Groves, N.; Dabrera, G.;  et~al.
\newblock {Effectiveness of Covid-19 Vaccines against the~B.1.617.2 (Delta)
  Variant}.
\newblock {\em N. Engl. J. Med.} {\bf 2021}, {\em
  385},~585--594.
\newblock {{https://doi.org/10.1056/NEJMoa2108891}}.

\bibitem[McCallum \em{et~al.}(2021)McCallum, Bassi, De~Marco, Chen, Walls,
  Di~Iulio, Tortorici, Navarro, Silacci-Fregni, Saliba,
  et~al.]{epsilon-vaccine-efficacy}
McCallum, M.; Bassi, J.; De~Marco, A.; Chen, A.; Walls, A.C.; Di~Iulio, J.;
  Tortorici, M.A.; Navarro, M.J.; Silacci-Fregni, C.; Saliba, C.;  et~al.
\newblock {SARS-CoV-2 immune evasion by variant B. 1.427/B. 1.429}.
\newblock {\em Science} {\bf 2021}, {\em 373},~648--654. https://doi.org/10.1126/science.abi7994.

\bibitem[Olliaro \em{et~al.}(2021)Olliaro, Torreele, and
  Vaillant]{vac_efficacy}
Olliaro, P.; Torreele, E.; Vaillant, M.
\newblock COVID-19 vaccine efficacy and effectiveness---The elephant (not) in
  the~room.
\newblock {\em  Lancet Microbe} {\bf 2021}, {\em 2},~279--2809.

\bibitem[Pormohammad \em{et~al.}(2021)Pormohammad, Zarei, Ghorbani, Mohammadi,
  Razizadeh, Turner, and~Turner]{vac_eff_mdpi}
Pormohammad, A.; Zarei, M.; Ghorbani, S.; Mohammadi, M.; Razizadeh, M.H.;
  Turner, D.L.; Turner, R.J.
\newblock Efficacy and Safety of COVID-19 Vaccines: A Systematic Review and
  Meta-Analysis of Randomized Clinical Trials.
\newblock {\em Vaccines} {\bf 2021}, {\em 9},~467.

\bibitem[Abdullahi \em{et~al.}(2020)Abdullahi, Onyango, Mukiira, Wamicwe,
  Githiomi, Kariuki, Mugambi, Wanjohi, Githuka, Nzioka, Orwa, Oronje, Kariuki,
  and Mayieka]{hand-washing}
Abdullahi, L.; Onyango, J.J.; Mukiira, C.; Wamicwe, J.; Githiomi, R.; Kariuki,
  D.; Mugambi, C.; Wanjohi, P.; Githuka, G.; Nzioka, C.;  et~al.
\newblock Community interventions in~Low—And Middle-Income Countries to
  inform COVID-19 control implementation decisions in~Kenya: A rapid systematic
  review.
\newblock {\em PLoS ONE} {\bf 2020}, {\em 15},~{e0242403}.
\newblock {{https://doi.org/10.1371/journal.pone.0242403}}.

\bibitem[Mukerjee \em{et~al.}(2021)Mukerjee, Chow, and~Li]{gondor}
Mukerjee, S.; Chow, C.M.; Li, M.
\newblock Mitigation strategies and compliance in~the~COVID-19 fight; how much
  compliance is enough?
\newblock {\em PLoS ONE} {\bf 2021}, {\em 16},~{e0239352}.
\newblock {{https://doi.org/10.1371/journal.pone.0239352}}.

\bibitem[Kashem \em{et~al.}(2021)Kashem, Baker, Gonz{\'a}lez, and
  Lee]{SCSsocialvulnerability}
Kashem, S.B.; Baker, D.M.; Gonz{\'a}lez, S.R.; Lee, C.A.
\newblock Exploring the~nexus between social vulnerability, built environment,
  and the~prevalence of COVID-19: A case study of Chicago.
\newblock {\em Sustain. Cities Soc.} {\bf 2021}, {\em 75},~103261.

\bibitem[Ugail \em{et~al.}(2021)Ugail, Aggarwal, Iglesias, Howard, Campuzano,
  Su{\'a}rez, Maqsood, Aadil, Mehmood, Gleghorn, et~al.]{SCSphysical-space-opt}
Ugail, H.; Aggarwal, R.; Iglesias, A.; Howard, N.; Campuzano, A.; Su{\'a}rez,
  P.; Maqsood, M.; Aadil, F.; Mehmood, I.; Gleghorn, S.;  et~al.
\newblock Social distancing enhanced automated optimal design of physical
  spaces in~the~wake of the~COVID-19 pandemic.
\newblock {\em Sustain. Cities Soc.} {\bf 2021}, {\em 68},~102791.

\bibitem[Ren \em{et~al.}(2021)Ren, Xi, Wang, Feng, Nasiri, Cao, and
  Haghighat]{SCSphysical-barrier}
Ren, C.; Xi, C.; Wang, J.; Feng, Z.; Nasiri, F.; Cao, S.J.; Haghighat, F.
\newblock Mitigating COVID-19 infection disease transmission in~indoor
  environment using physical barriers.
\newblock {\em Sustain. Cities Soc.} {\bf 2021}, {\em 74},~103175.

\bibitem[Srivastava \em{et~al.}(2021)Srivastava, Zhao, Manay, and
  Chen]{SCSoffice-threat}
Srivastava, S.; Zhao, X.; Manay, A.; Chen, Q.
\newblock Effective ventilation and air disinfection system for reducing
  Coronavirus Disease 2019 (COVID-19) infection risk in~office buildings.
\newblock {\em Sustain. Cities Soc.} {\bf 2021}, {\em {75}}, 103408.

\bibitem[Grekousis and Liu(2021)]{SCSreview-contacttracing}
Grekousis, G.; Liu, Y.
\newblock Digital contact tracing, community uptake, and~proximity awareness
  technology to fight COVID-19: A systematic review.
\newblock {\em Sustain. Cities Soc.} {\bf 2021}, {\em 71},~102995.

\bibitem[Bian \em{et~al.}(2020)Bian, Zhou, Bello, and
  Lukowicz]{bian2020wearable}
Bian, S.; Zhou, B.; Bello, H.; Lukowicz, P.
\newblock A wearable magnetic field based proximity sensing system for
  monitoring COVID-19 social distancing.
\newblock In Proceedings of the~2020 International Symposium
  on Wearable Computers, {Virtual Event, 12--17 September} 2020; pp. 22--26.

\bibitem[Fazio \em{et~al.}(2020)Fazio, Buzachis, Galletta, Celesti, and
  Villari]{bluetooth2020}
Fazio, M.; Buzachis, A.; Galletta, A.; Celesti, A.; Villari, M.
\newblock {A proximity-based indoor navigation system tackling the~COVID-19
  social distancing measures}.
\newblock In Proceedings of the IEEE Symposium on Computers and Communications, {Rennes, France, 7--10 July} 2020.
\newblock {{https://doi.org/10.1109/ISCC50000.2020.9219634}}.

\bibitem[Shafiee \em{et~al.}(2017)Shafiee, Chywl, Li, and~Wong]{yolov2}
Shafiee, M.J.; Chywl, B.; Li, F.; Wong, A.
\newblock Fast YOLO: A fast you only look once system for real-time embedded
  object detection in~video.
\newblock {\em arXiv} {\bf 2017}, arXiv:1709.05943.

\bibitem[{Wojke} \em{et~al.}(2017){Wojke}, {Bewley}, and
  {Paulus}]{deepsort-2017}
{Wojke}, N.; {Bewley}, A.; {Paulus}, D.
\newblock Simple online and realtime tracking with a~deep association metric.
\newblock In Proceedings of the~2017 IEEE International Conference on Image
  Processing (ICIP), {Beijing, China, 17--20 September} 2017; pp. 3645--3649.
\newblock {{https://doi.org/10.1109/ICIP.2017.8296962}}.

\bibitem[Chen \em{et~al.}(2019)Chen et~al.]{imageNvision-license-plate}
Chen, R.C.
\newblock Automatic License Plate Recognition via sliding-window darknet-YOLO
  deep learning.
\newblock {\em Image Vis. Comput.} {\bf 2019}, {\em 87},~47--56.

\bibitem[Ye \em{et~al.}(2020)Ye, Hong, Chen, Hsiao, and
  Fu]{imageNvision-road-marking}
Ye, X.Y.; Hong, D.S.; Chen, H.H.; Hsiao, P.Y.; Fu, L.C.
\newblock A two-stage real-time YOLOv2-based road marking detector with
  lightweight spatial transformation-invariant classification.
\newblock {\em Image Vis. Comput.} {\bf 2020}, {\em 102},~103978.

\bibitem[Chiang \em{et~al.}(2021)Chiang, Wang, and
  Chen]{imageNvision-pedestrian-detection}
Chiang, S.H.; Wang, T.; Chen, Y.F.
\newblock Efficient pedestrian detection in~top-view fisheye images using
  compositions of perspective view patches.
\newblock {\em Image Vis. Comput.} {\bf 2021}, {\em 105},~104069.
\newblock {{https://doi.org/10.1016/j.imavis.2020.104069}}.

\bibitem[Wu \em{et~al.}(2020)Wu, Lv, Jiang, and
  Song]{imageNvision-apple-flower}
Wu, D.; Lv, S.; Jiang, M.; Song, H.
\newblock Using channel pruning-based YOLO v4 deep learning algorithm for the
  real-time and accurate detection of apple flowers in~natural environments.
\newblock {\em Comput. Electron. Agric.} {\bf 2020}, {\em
  178},~105742.

\bibitem[Ansari and Singh(2021)]{ansari2021monitoring}
Ansari, M.A.; Singh, D.K.
\newblock Monitoring social distancing through human detection for
  preventing/reducing COVID spread.
\newblock {\em Int. J. Inf. Technol.} {\bf 2021}, {\em
  {13}}, {1255--1264}.

\bibitem[Ahmed \em{et~al.}(2021{\natexlab{a}})Ahmed, Ahmad, and
  Jeon]{SCStopviewDL}
Ahmed, I.; Ahmad, M.; Jeon, G.
\newblock Social distance monitoring framework using deep learning architecture
  to control infection transmission of COVID-19 pandemic.
\newblock {\em Sustain. Cities Soc.} {\bf 2021}, {\em 69},~102777.

\bibitem[Ahmed \em{et~al.}(2021{\natexlab{b}})Ahmed, Ahmad, Rodrigues, Jeon,
  and Din]{ahmed2021deep}
Ahmed, I.; Ahmad, M.; Rodrigues, J.J.; Jeon, G.; Din, S.
\newblock A deep learning-based social distance monitoring framework for
  COVID-19.
\newblock {\em Sustain. Cities Soc.} {\bf 2021}, {\em 65},~102571.

\bibitem[Qin and Xu(2021)]{qin2021reaserch}
Qin, J.; Xu, N.
\newblock Reaserch and implementation of social distancing monitoring
  technology based on SSD.
\newblock {\em Procedia Comput. Sci.} {\bf 2021}, {\em 183},~768--775.

\bibitem[Rahim \em{et~al.}(2021)Rahim, Maqbool, and~Rana]{rahim2021monitoring}
Rahim, A.; Maqbool, A.; Rana, T.
\newblock Monitoring social distancing under various low light conditions with
  deep learning and a~single motionless time of flight camera.
\newblock {\em PLoS ONE} {\bf 2021}, {\em 16},~e0247440.

\bibitem[Su \em{et~al.}(2021)Su, He, Qing, Niu, Cheng, and~Peng]{su2021novel}
Su, J.; He, X.; Qing, L.; Niu, T.; Cheng, Y.; Peng, Y.
\newblock A novel social distancing analysis in~urban public space: A new
  online spatio-temporal trajectory approach.
\newblock {\em Sustain. Cities Soc.} {\bf 2021}, {\em 68},~102765.

\bibitem[Rezaei and Azarmi(2020)]{deepsocial-mdpi-applied-sci}
Rezaei, M.; Azarmi, M.
\newblock {Deepsocial: Social distancing monitoring and infection risk
  assessment in~covid-19 pandemic}.
\newblock {\em Appl. Sci.} {\bf 2020}, {\em 10},~1--29.
\newblock {{https://doi.org/10.3390/app10217514}}.

\bibitem[Yang \em{et~al.}(2021)Yang, Yurtsever, Renganathan, Redmill, and
  {\"O}zg{\"u}ner]{vision-based-social-distancing-critical-density}
Yang, D.; Yurtsever, E.; Renganathan, V.; Redmill, K.A.; {\"O}zg{\"u}ner,
  {\"U}.
\newblock A vision-based social distancing and critical density detection
  system for COVID-19.
\newblock {\em Sensors} {\bf 2021}, {\em 21},~4608.

\bibitem[Punn \em{et~al.}(2020)Punn, Sonbhadra, and
  Agarwal]{social-distancing-fine-tuned-yolo-deepsort}
Punn, N.S.; Sonbhadra, S.K.; Agarwal, S.
\newblock {Monitoring COVID-19 social distancing with person detection and
  tracking via fine-tuned YOLO v3 and Deepsort techniques}. {\em {arXiv}} {\bf 2020}, {arXiv:2005.01385}.

\bibitem[Eikenberry \em{et~al.}(2020)Eikenberry, Mancuso, Iboi, Phan,
  Eikenberry, Kuang, Kostelich, and~Gumel]{wearing-masks}
Eikenberry, S.E.; Mancuso, M.; Iboi, E.; Phan, T.; Eikenberry, K.; Kuang, Y.;
  Kostelich, E.; Gumel, A.B.
\newblock To mask or not to mask: Modeling the~potential for face mask use by
  the~general public to curtail the~COVID-19 pandemic.
\newblock {\em Infect. Dis. Model.} {\bf 2020}, {\em 5},~293--308.

\bibitem[Kampf(2020)]{touch-spread-1}
Kampf, G.
\newblock Potential role of inanimate surfaces for the~spread of coronaviruses
  and their inactivation with disinfectant agents.
\newblock {\em Infect. Prev. Pract.} {\bf 2020}, {\em 2},~100044.

\bibitem[Warnes \em{et~al.}(2015)Warnes, Little, and~Keevil]{touch-spread-2}
Warnes, S.L.; Little, Z.R.; Keevil, C.W.
\newblock Human coronavirus 229E remains infectious on common touch surface
  materials.
\newblock {\em MBio} {\bf 2015}, {\em 6}, {e01697-15}.

\bibitem[Roy \em{et~al.}(2020)Roy, Nandy, Ghosh, Dutta, Biswas, and
  Das]{Moxa3k}
Roy, B.; Nandy, S.; Ghosh, D.; Dutta, D.; Biswas, P.; Das, T.
\newblock Moxa: A deep learning based unmanned approach for real-time
  monitoring of people wearing medical masks.
\newblock {\em Trans. Indian Natl. Acad. Eng.} {\bf
  2020}, {\em 5},~509--518.

\bibitem[Mohan \em{et~al.}(2021)Mohan, Paul, and~Chirania]{TinyMask}
Mohan, P.; Paul, A.J.; Chirania, A.
\newblock A tiny CNN architecture for medical face mask detection for
  resource-constrained endpoints. In~{\em Innovations in~Electrical and
  Electronic Engineering}; Springer: {Berlin/Heidelberg, Germany,} 
  2021; pp. 657--670.

\bibitem[Loey \em{et~al.}(2021)Loey, Manogaran, Taha, and
  Khalifa]{SCSmask-detection}
Loey, M.; Manogaran, G.; Taha, M.H.N.; Khalifa, N.E.M.
\newblock Fighting against COVID-19: A novel deep learning model based on
  YOLO-v2 with ResNet-50 for medical face mask detection.
\newblock {\em Sustain. Cities Soc.} {\bf 2021}, {\em 65},~102600.

\bibitem[Wu \em{et~al.}(2022)Wu, Li, Zeng, and~Li]{imageNvision-facemask}
Wu, P.; Li, H.; Zeng, N.; Li, F.
\newblock FMD-Yolo: An efficient face mask detection method for COVID-19
  prevention and control in~public.
\newblock {\em Image Vis. Comput.} {\bf 2022}, {\em 117},~104341.
\newblock {https://doi.org/10.1016/j.imavis.2021.104341}.

\bibitem[Hassan \em{et~al.}(2021)Hassan, Sritharan, Jayatilaka, Godaliyadda,
  Ekanayake, Herath, and~Ekanayake]{HandsOff}
Hassan, A.; Sritharan, S.; Jayatilaka, G.; Godaliyadda, R.I.; Ekanayake, P.B.;
  Herath, V.; Ekanayake, J.B.
\newblock Hands Off: A Handshake Interaction Detection and Localization Model
  for COVID-19 Threat Control.
\newblock {\em arXiv} {\bf 2021}, arXiv:2110.09571.

\bibitem[Shinde \em{et~al.}(2018)Shinde, Kothari, and~Gupta]{shinde2018yolo}
Shinde, S.; Kothari, A.; Gupta, V.
\newblock YOLO based human action recognition and localization.
\newblock {\em Procedia Comput. Sci.} {\bf 2018}, {\em 133},~831--838.

\bibitem[Sefidgar \em{et~al.}(2015)Sefidgar, Vahdat, Se, and
  Mori]{sefidgar2015discriminative}
Sefidgar, Y.S.; Vahdat, A.; Se, S.; Mori, G.
\newblock Discriminative key-component models for interaction detection and
  recognition.
\newblock {\em Comput. Vis. Image Underst.} {\bf 2015}, {\em
  135},~16--30.

\bibitem[Van~Gemeren \em{et~al.}(2018)Van~Gemeren, Poppe, and
  Veltkamp]{van2018hands}
Van~Gemeren, C.; Poppe, R.; Veltkamp, R.C.
\newblock Hands-on: Deformable pose and motion models for spatiotemporal
  localization of fine-grained dyadic interactions.
\newblock {\em EURASIP J. Image Video Process.} {\bf 2018}, {\em
  2018},~{16}.

\bibitem[Jones \em{et~al.}(2020)Jones, Qureshi, Temple, Larwood, Greenhalgh,
  and Bourouiba]{jones2020two}
Jones, N.R.; Qureshi, Z.U.; Temple, R.J.; Larwood, J.P.; Greenhalgh, T.;
  Bourouiba, L.
\newblock Two metres or one: What is the~evidence for physical distancing in
  covid-19?
\newblock {\em BMJ} {\bf 2020}, {\em 370}. {https://doi.org/10.1136/bmj.m3223.}

\bibitem[Kwon \em{et~al.}(2021)Kwon, Joshi, Lo, Drew, Nguyen, Guo, Ma, Mehta,
  Shebl, Warner, et~al.]{kwon2021association}
Kwon, S.; Joshi, A.D.; Lo, C.H.; Drew, D.A.; Nguyen, L.H.; Guo, C.G.; Ma, W.;
  Mehta, R.S.; Shebl, F.M.; Warner, E.T.;  et~al.
\newblock Association of social distancing and face mask use with risk of
  COVID-19.
\newblock {\em Nat. Commun.} {\bf 2021}, {\em 12},~{3737}.

\bibitem[Durkin \em{et~al.}(2021)Durkin, Jackson, and~Usher]{durkin2021touch}
Durkin, J.; Jackson, D.; Usher, K.
\newblock Touch in~times of COVID-19: Touch hunger hurts.
\newblock {\em J. Clin. Nurs.} {\bf 2021}, \emph{{30}}, {e4--e5}. 

\bibitem[Qian and Jiang(2020)]{qian2020covid}
Qian, M.; Jiang, J.
\newblock COVID-19 and social distancing.
\newblock {\em J. Public Health} {\bf 2020}, {\em {30}}, {259--261}.

\bibitem[Verani \em{et~al.}(2020)Verani, Clodfelter, Menon, Chevinsky, Victory,
  and Hakim]{verani2020social}
Verani, A.; Clodfelter, C.; Menon, A.N.; Chevinsky, J.; Victory, K.; Hakim, A.
\newblock Social distancing policies in~22 African countries during the
  COVID-19 pandemic: A desk review.
\newblock {\em  Pan Afr. Med J.} {\bf 2020}, {\em 37}, {46}.

\bibitem[Bochkovskiy \em{et~al.}(2020)Bochkovskiy, Wang, and~Liao]{yolov4}
Bochkovskiy, A.; Wang, C.Y.; Liao, H.Y.M.
\newblock YOLOv4: Optimal Speed and Accuracy of Object Detection. {\em {arXiv}} {\bf 2020}, {arXiv:2004.10934}.

\bibitem[Canny(1986)]{canny}
Canny, J.
\newblock A computational approach to edge detection.
\newblock {\em IEEE Trans. Pattern Anal. Mach. Intell.}
  {\bf 1986}, {\em {PAMI-8}}, 679--698.

\bibitem[Lin \em{et~al.}(2014)Lin, Maire, Belongie, Hays, Perona, Ramanan,
  Doll{\'a}r, and~Zitnick]{COCO}
Lin, T.Y.; Maire, M.; Belongie, S.; Hays, J.; Perona, P.; Ramanan, D.;
  Doll{\'a}r, P.; Zitnick, C.L.
\newblock Microsoft coco: Common objects in~context.
\newblock In {\em Proceedings of the~European Conference on Computer Vision};
  Springer: {Berlin/Heidelberg, Germany,} 
  2014; pp. 740--755.

\bibitem[Dendorfer \em{et~al.}(2020)Dendorfer, Rezatofighi, Milan, Shi,
  Cremers, Reid, Roth, Schindler, and~Leal-Taix{\'e}]{mot-dataset-2020}
Dendorfer, P.; Rezatofighi, H.; Milan, A.; Shi, J.; Cremers, D.; Reid, I.;
  Roth, S.; Schindler, K.; Leal-Taix{\'e}, L.
\newblock Mot20: A benchmark for multi object tracking in~crowded scenes.
\newblock {\em arXiv} {\bf 2020}, arXiv:2003.09003.

\bibitem[Jiang \em{et~al.}(2010)Jiang, Gao, and~Xu]{perspective-transform}
Jiang, Y.; Gao, F.; Xu, G.
\newblock Computer vision-based multiple-lane detection on straight road and in
  a~curve.
\newblock In Proceedings of the~2010 International Conference on Image Analysis
  and Signal Processing, {Zhejiang, China, 9--11 April} 2010; pp. 114--117.

\bibitem[Von~Luxburg(2007)]{spectral-clustering}
Von~Luxburg, U.
\newblock A tutorial on spectral clustering.
\newblock {\em Stat. Comput.} {\bf 2007}, {\em 17},~395--416.

\bibitem[Milan \em{et~al.}(2016)Milan, Leal-Taix{\'e}, Reid, Roth, and
  Schindler]{mot-dataset-2016}
Milan, A.; Leal-Taix{\'e}, L.; Reid, I.; Roth, S.; Schindler, K.
\newblock MOT16: A benchmark for multi-object tracking.
\newblock {\em arXiv} {\bf 2016}, arXiv:1603.00831.

\bibitem[Leal-Taix{\'e} \em{et~al.}(2015)Leal-Taix{\'e}, Milan, Reid, Roth, and
  Schindler]{mot-dataset-2015}
Leal-Taix{\'e}, L.; Milan, A.; Reid, I.; Roth, S.; Schindler, K.
\newblock Motchallenge 2015: Towards a~benchmark for multi-target tracking.
\newblock {\em arXiv} {\bf 2015}, arXiv:1504.01942.

\bibitem[Ryoo and Aggarwal(2010)]{UT-Interaction-Data}
Ryoo, M.S.; Aggarwal, J.K.
\newblock {UT}-{I}nteraction {D}ataset, {ICPR} contest on {S}emantic
  {D}escription of {H}uman {A}ctivities ({SDHA}). 2010. 
\newblock Available online: \url{http://cvrc.ece.utexas.edu/SDHA2010/Human\_Interaction.html}  {(accessed on}).  


\bibitem[Ryoo and Aggarwal(2009)]{UT2}
Ryoo, M.S.; Aggarwal, J.K.
\newblock Spatio-Temporal Relationship Match: Video Structure Comparison for
  Recognition of Complex Human Activities.
\newblock In Proceedings of the~IEEE International Conference on Computer
  Vision (ICCV),  {Kyoto, Japan, 29 September--2 October} 2009.

\bibitem[Everingham \em{et~al.}(2010)Everingham, Van~Gool, Williams, Winn, and
  Zisserman]{pascalvoc}
Everingham, M.; Van~Gool, L.; Williams, C.K.; Winn, J.; Zisserman, A.
\newblock The pascal visual object classes (voc) challenge.
\newblock {\em Int. J. Comput. Vis.} {\bf 2010}, {\em
  88},~303--338.

\bibitem[Kuznetsova \em{et~al.}(2018)Kuznetsova, Rom, Alldrin, Uijlings,
  Krasin, Pont-Tuset, Kamali, Popov, Malloci, Kolesnikov, et~al.]{openimages}
Kuznetsova, A.; Rom, H.; Alldrin, N.; Uijlings, J.; Krasin, I.; Pont-Tuset, J.;
  Kamali, S.; Popov, S.; Malloci, M.; Kolesnikov, A.;  et~al.
\newblock The open images dataset v4: Unified image classification, object
  detection, and~visual relationship detection at scale.
\newblock {\em arXiv} {\bf 2018}, arXiv:1811.00982.

\bibitem[Robertson(2008)]{map-paper}
Robertson, S.
\newblock A new interpretation of average precision.
\newblock In Proceedings of the~31st Annual International
  ACM SIGIR Conference on Research and Development in~Information Retrieval, {Singapore, 20--24 July} 
  2008; pp. 689--690.

\bibitem[{Padilla} \em{et~al.}(2020){Padilla}, {Netto}, and~{da
  Silva}]{padilla2}
{Padilla}, R.; {Netto}, S.L.; {da Silva}, E.A.B.
\newblock A Survey on Performance Metrics for Object-Detection Algorithms.
\newblock In Proceedings of the~2020 International Conference on Systems,
  Signals and Image Processing (IWSSIP), {Niteroi, Brazil, 1--3 July} 2020; pp. 237--242.

\end{thebibliography}
\end{document}